\documentclass[10pt,journal,compsoc]{IEEEtran}
\usepackage{comment}
\usepackage{ifpdf}
\usepackage{ragged2e}

\usepackage{tikz}

\usepackage[colorlinks, linkcolor=hotpink, anchorcolor=blue, citecolor=hotpink, ]{hyperref}

\ifCLASSOPTIONcompsoc
  \usepackage[nocompress]{cite}
\else
  \usepackage{cite}
\fi
\ifCLASSINFOpdf
\else
\fi

\usepackage{amsmath,amssymb,amsfonts}
\usepackage{amsthm}
\usepackage{algorithmic}
\usepackage{graphicx}
\usepackage{textcomp}
\usepackage{xcolor}
\usepackage{multirow}
\usepackage{booktabs} 
\usepackage{threeparttable}
\usepackage{array}
\usepackage{mdwmath}
\usepackage{mdwtab}
\usepackage{hyperref}
\usepackage{mathrsfs}
\usepackage{threeparttable}
\usepackage[switch]{lineno}
\hypersetup{
    colorlinks = true,
    linkcolor = black,
    anchorcolor = black,
    citecolor = black,
    filecolor = black,
    urlcolor = black
    }

\newtheorem{theorem}{Theorem}

\newtheorem{definition}{Definition}

\usepackage[numbers]{natbib}
\usepackage[ruled,boxed,commentsnumbered]{algorithm2e}

\newcommand{\INDSTATE}[1][1]{\STATE\hspace{#1\algorithmicindent}}
\usepackage{makecell}
\usepackage{url}

 \usepackage{comment}

\ifCLASSOPTIONcompsoc
  \usepackage[nocompress]{cite}
\else
  \usepackage{cite}
\fi
\ifCLASSINFOpdf
\else
\fi

\begin{document}
%
\title{Foundations and Frontiers of Graph Learning Theory}
%
%
%
%

\author{Yu~Huang, 
        Min~Zhou,
        Menglin~Yang, 
        Zhen~Wang, 
        Muhan~Zhang, 
        Jie~Wang, 
        Hong~Xie, 
        Hao~Wang,
        Defu~Lian,
        and~Enhong~Chen,~\IEEEmembership{Fellow,~IEEE}
\IEEEcompsocitemizethanks{\IEEEcompsocthanksitem Y. Huang, J. Wang, H. Xie, H. Wang, D. Lian and E. Chen are with the University of Science and Technology of China, Hefei, Anhui, 230027.
E-mail:hy123123@mail.ustc.edu.cn, \{jiewangx, hongx87, wanghao3, liandefu, cheneh\}@ustc.edu.cn 
\IEEEcompsocthanksitem M. Zhou is with Huawei and the work is independent to the position or resource in the company. E-mail: zhoum1900@163.com. \protect
\IEEEcompsocthanksitem M. Yang is with Chinese University of Hong Kong. E-mail: menglin.yang@outlook.com\protect
\IEEEcompsocthanksitem Z. Wang is with Sun Yat-sen University. E-mail:joneswong.ml@gmail.com\protect
\IEEEcompsocthanksitem M. Zhang is with Peking University. E-mail: muhan@pku.edu.cn\protect\\
}
}

%
%


\markboth{Journal of \LaTeX\ Class Files}%
{Shell \MakeLowercase{\textit{et al.}}: Bare Demo of IEEEtran.cls for Computer Society Journals}
%



\IEEEtitleabstractindextext{%
\begin{abstract}
Recent advancements in graph learning have revolutionized the way to understand and analyze data with complex structures. Notably, Graph Neural Networks (GNNs), i.e. neural network architectures designed for learning graph representations, have become a popular paradigm. With these models being usually characterized by intuition-driven design or highly intricate components, placing them within the theoretical analysis framework to distill the core concepts, helps understand the key principles that drive the functionality better and guide further development. Given this surge in interest, this article provides a comprehensive summary of the theoretical foundations and breakthroughs concerning the approximation and learning behaviors intrinsic to prevalent graph learning models. Encompassing discussions on fundamental aspects such as 
expressiveness power, generalization, optimization, and unique phenomena such as over-smoothing and over-squashing, this piece delves into the theoretical foundations and frontier driving the evolution of graph learning. In addition, this article also presents several challenges and further initiates discussions on possible solutions.

\end{abstract}

}

\maketitle

\IEEEdisplaynontitleabstractindextext

%
\IEEEpeerreviewmaketitle

\IEEEraisesectionheading{\section{Introduction}\label{sec:introduction}}

%
%
%
%

\IEEEPARstart{R}eal-world datasets can be naturally represented as graphs, where nodes represent entities interconnected by edges denoting relationships. 
Graph related tasks encompass a broad spectrum, spanning node classification~\cite{yan2024rethinking,li2024sefraud,shi2023label}, link prediction~\cite{li2022bsal,yang2022hrcf}, graph classification/regression~\cite{zhang2018end,liu2023mata}, as well as generation tasks~\cite{guo2022systematic,jin2022antibody}. The applications of these tasks extend to diverse areas such as property prediction in molecules~\cite{zhang2021motif,gengnovo}, traffic analysis~\cite{guo2020optimized,deng2022transposed}, social network analysis~\cite{deng2022markov, wang2019mcne}, physics simulations~\cite{sanchez2020learning,han2022learning}, and combinatorial optimization~\cite{wang2023learning,kuang2024towards}.

Solving these diverse tasks demands a sufficiently rich embedding of the graph or node that captures
structural properties as well as attribute information. While graph embeddings have been a widely-studied topic, including spectral embeddings and graph kernels, recently, Graph Neural Networks
(GNNs) have emerged as an empirically and broadly successful model class that, as opposed to, e.g., spectral embeddings, allows to adapt the embedding to the task at hand, generalizes to other graphs of the same input type, and incorporates attributes. 

The objective of graph learning entails the discovery of a function that can approximate the target function based on the available information of a given graph. This process involves several sequential steps. Firstly, it identifies a comprehensive set of functions, such as graph neural networks, capable of representing the target function with sufficient expressiveness. Subsequently, the function that provides the best approximation of the target function is estimated by minimizing a designated loss function (such as noise contrastive estimation loss or cross-entropy). Finally, the estimated function from the previous step is utilized to predict outcomes on test data, resulting test errors that composed of error accumulated in the above steps. In crafting an optimal graph model, three pivotal considerations typically shape the approach: 

\begin{itemize}
    \item \textbf{Expressiveness} also known as representation power explores if target functions can be approximated well by a graph model. For functions on graphs, representational power has mainly been studied in terms of graph isomorphism, i.e., studying the graphs that a designed model can distinguish or not.

    Topics to this question include graph isomorphism testing, subgraph counting,  representing invariance/equivariance under permutations, etc.

\item \textbf{Generalization} asks how well the estimated function is performing according to the population risk, as a function of the number of data points and model properties. To quantify the generalization ability, the generalization error bound is a good measurement. Typically, generalization analyses involve the complexity of the model class, the target function, the data, and the optimization procedure. 
\item \textbf{Optimization} investigates conditions under which training algorithms like gradient descent or training tricks can be shown to be provably effective in terms of faster convergence or good generalization of a given GNN. Possible solutions are training tricks or strategies to ensure the algorithm converges to the global or acceptable local minima as soon as possible.

\end{itemize}

By carefully addressing one or more of the aforementioned aspects mentioned-above, considerable graph learning models have emerged daily, as consolidated in references~\cite{goyal2018graph,wu2020comprehensive,li2024guest}. Since the models evolve to be characterized by highly intricate network architectures to meet the diverse scenario and high performance needs, the importance of comprehending the fundamental principles underlying these models becomes evident. 
Regarding the rapid growth of theoretical analysis on graph models, there are fragments in the area~\cite{jegelka2022theory,morris2021weisfeiler,zhang2023rethinking}, with overlap in certain subjects covered but different focuses.  A holistic landscape of the progress and advancements in graph learning is still vacant. 

For instance, Jegelka \cite{jegelka2022theory} summarizes a selection of emerging theoretical results on approximation and generalization properties of messaging passing GNNs, with emphasis on the mathematical connections, serving for the mathematicians.  Regarding expressiveness, Morris et al.~\cite{morris2021weisfeiler} provide an extensive review on algorithms and neural architectures based on the Weisfeiler-Leman algorithm, a well-known heuristic for the graph isomorphism problem. Zhang et al.~\cite{zhang2023rethinking} further broaden the regime to popular GNN variants through the lens of graph bi-connectivity. In the domain of deep learning, training strategies and hyper-parameter selections play crucial roles. However, current surveys on optimization in deep learning primarily focus on general machine learning or neural networks operating with Euclidean data \cite{sun2019optimization,suh2024survey}. There is a clear need for a comprehensive summary of optimization strategies tailored specifically for graph structures to enhance the performance and efficiency of graph neural networks and related models. Very recently, Morris et al.~\cite{morrisposition} spotlight the future directions in the theory of graph machine learning that is most related to this article while the authors only provide a brief summary of existing works.

To fill the gap, this article delves into the aforementioned three aspects by giving precise mathematical settings and summarizing theoretical tools as well as current progress, offering readers resource on the design principles in the context of graph-structured data. 
In practical applications, the ideal graph models exhibit strong expressive power, minimal generalization error, as well as rapid and stable convergence speed.  In theoretical analysis, however, the three subjects are often explored independently to derive possible results or insights. For each topic, we first explain their respective goals and basic concepts, then provide a detailed classification of the methods used with relevant theoretical results. Finally, we establish their interconnections, suggesting potential research directions for future graph learning theories.  In addition to common fundamental aspects shared by typical graph models, issues such as performance degradation in deeper Graph Neural Networks (GNNs) and the information bottleneck stemming from long-range dependencies, known as over-squashing, are critical phenomena. Addressing and mitigating these challenges are pivotal for improving the efficacy and robustness of GNNs, particularly in tasks necessitating intricate hierarchical representations of graph-structured data. Given that resolve these two issues involves a multifaceted approach, they are detailed in a separate section for a comprehensive review.

Different from previous surveys focusing more on mathematical connections, we aim to summarize key principles that drive the functionality and address the unique challenges and requirements posed by graphs in machine learning tasks. We will try to avoid too many technicalities and make the survey serving for both theorists as well practitioners in various fields. The reminder of this work unfolds as follows: Section~\ref{sec:pre} gives a summary of fundamental methods in graph learning. Section~\ref{sec:exp_power} to Section~\ref{sec:opt} details diverse theoretical analyses and results of expressive power, generalization, and optimization, respectively. The formulation and solutions of over-smoothing and over-squashing are documented in Section~\ref{sec:exp_long_range}. The paper concludes in Section~\ref{sec:conclusion}.
\section{Preliminary}
\label{sec:pre}
Graph embedding, graph kernels, and GNNs are fundamental approaches for representing and analyzing graph-structured data. While graph embedding and graph kernels have been effective in representing and analyzing graph-structured data, they face challenges in capturing complex graph interactions and may require pre-processing steps. GNNs address these limitations by combining the power of neural networks with the expressive capacity of graphs, enabling end-to-end learning of node and graph representations. Recently, Graph Transformers have emerged as an advanced technique in graph learning, applying self-attention mechanisms to capture long-range dependencies between nodes while incorporating graph structural information. These advancements have opened up new possibilities for understanding graph-structured data in various domains. The following subsections provide a detailed overview of these approaches.

\subsection{Graph Embedding and Graph Kernels}
Graph embedding and graph kernels are two fundamental approaches for representing and analyzing graph-structured data. Graph embedding converts an input graph $\mathcal{G} = (V, E)$ into a low-dimensional vector space, capturing essential graph properties such as connectivity, community structures, or clustering patterns. Graph embedding methods can be categorized into matrix factorization-based methods (e.g., GF~\cite{ahmed2013distributed}, GraRep~\cite{cao2015grarep}), random walk-based methods (e.g., DeepWalk~\cite{perozzi2014deepwalk}, node2vec~\cite{grover2016node2vec}), and deep learning-based methods (e.g., SDNE~\cite{wang2016structural}, DNGR~\cite{cao2016deep}). While effective in capturing various graph properties, graph embedding techniques may not fully capture complex interactions and dependencies in graphs.

Graph kernels, a subset of kernel functions, quantify the similarity between pairs of graphs by computing the inner product between their feature representations in a high-dimensional space. The Weisfeiler-Lehman (WL) subtree kernel~\cite{shervashidze2011weisfeiler} has been highly influential, inspiring extensions such as the WL Optimal Assignment kernel~\cite{kriege2016valid} and the WL shortest-path kernel~\cite{shervashidze2011weisfeiler}. Shortest-path kernels~\cite{borgwardt2005shortest} measure similarity based on the properties of shortest paths, considering both path lengths and vertex attributes. Random walk kernels~\cite{gartner2003graph,kashima2003marginalized} assess similarity by comparing label sequences of random walks on graphs. Graph kernels provide a comprehensive measure of graph similarity by considering both structure and label information, making them suitable for various graph comparison tasks.

Despite their strengths, graph embedding and graph kernels face challenges in capturing complex graph interactions and may require pre-processing steps. GNNs address these limitations by combining the power of neural networks with the expressive capacity of graphs, enabling end-to-end learning of node and graph representations.

\subsection{Graph Neural Networks}
GNNs have emerged as a powerful framework for analyzing graph-structured data, leveraging the message-passing mechanism to enable nodes to iteratively update their representations by aggregating information from their neighbors. GNNs can be broadly categorized into three types: spectral GNNs, spatial GNNs, and geometric GNNs based on the way they operate on graph-structured data.

\textbf{Spectral GNNs} operate on the spectral representation of the graph, which is obtained by the eigendecomposition of the graph Laplacian matrix~\cite{shuman2013emerging, sandryhaila2013discrete}. Spectral convolutions are defined in the Fourier domain and the graph Fourier transform of a signal $x$ is given by $\mathcal{F}(x)=\mathbf{U}^T x$, and the inverse transform is $\mathcal{F}^{-1}(\hat{x})=\mathbf{U} \hat{x}$. 
Then, the graph convolution of $x$ with a filter $\mathbf{g}$ is defined as:
\begin{equation}
x *_\mathcal{G} \mathbf{g} = \mathbf{U}\left(\mathbf{U}^T x \odot \mathbf{U}^T \mathbf{g}\right),
\end{equation}
where $\odot$ denotes element-wise multiplication.

Several spectral GNN variants have been proposed, including Spectral CNN \cite{bruna2013spectral}, ChebNet \cite{defferrard2016convolutional}, and GCN \cite{kipf2017semi}. However, spectral GNNs face challenges in generalizing across different graph structures and suffer from high computational complexity.

\textbf{Spatial GNNs} operate directly on the graph structure, leveraging spatial relationships between nodes to perform graph convolutions using a message-passing mechanism. 

\textit{The Message Passing Neural Network (MPNN)}~\cite{gilmer2017neural} provides a general framework for spatial GNNs, which is defined as:
\begin{equation}
x_v^{(k)} = U_k(x_v^{(k-1)}, \sum_{u\in \mathcal{N}(v)} M_k(x_v^{(k-1)}, x_u^{(k-1)}, x^e_{vu})).
\end{equation}
where $x_v^{(k)}$ represents the embedding of node $v$ at layer $k$, $\mathcal{N}(v)$ denotes the neighborhood of node $v$, $M_k(\cdot)$ is the message function that computes the message from node $u$ to node $v$, $U_k(\cdot)$ is the update function that updates the node embedding based on the aggregated messages, and $\mathbf{x}^e_{vu}$ represents the edge features between nodes $v$ and $u$ (if available).

\textit{GraphSAGE}~\cite{hamilton2017inductive} addresses the inductive and scalability by employing neighborhood sampling:
\begin{equation}
    x^{(k)}_v=\phi(\mathbf{W}^{(k)}\cdot \mathrm{AGG}(\{x_u^{(k-1)}, \forall u \in S_{\mathcal{N}(v)}\})),
\end{equation}
where $S_{\mathcal{N}(v)}$ represents the sampled neighborhood of node $v$, $\mathrm{AGG}(\cdot)$ is the aggregation function, and $\phi(\cdot)$ is an activation function.

Graph Attention Network (GAT) \cite{velickovic2017graph} introduces attention mechanisms to learn the relative importance of neighboring nodes:
\begin{equation}
    x_v^{(k)} = \phi(\sum_{u\in\mathcal{N}(v)\cup v}\alpha_{vu}^{(k)}\mathbf{W}^{(k)}x_u^{(k-1)}),
\end{equation}
where $\alpha_{vu}^{(k)}$ represents the attention weight assigned to the edge between nodes $v$ and $u$ at layer $k$.

Graph Isomorphism Network (GIN) \cite{xu2019how} introduces an adjustable weight parameter to better capture structural information:
\begin{equation}
    x_v^{(k)} = \mathrm{MLP}((1+\epsilon^{(k)})x_v^{(k-1)}+\sum_{u\in \mathcal{N}(v)}x_u^{(k-1)}),
\end{equation}
where $\epsilon^{(k)}$ is a learnable parameter that adjusts the weight of the central node's own features.

\textbf{Geometric GNNs} operate on graphs with additional geometric features, such as node positions and edge lengths or angles, to capture the spatial relationships between nodes and edges in graph-structured data. 
These additional geometric features are particularly relevant in domains like molecular modeling, where the spatial arrangement of atoms and bonds plays a crucial role in determining the properties and behavior of molecules. By leveraging the geometric information, Geometric GNNs can learn more expressive and informative representations compared to standard GNNs. However, when dealing with geometric graphs, Geometric GNNs must also consider the symmetries and equivariances present in the data. 

To address these challenges, several Geometric GNN architectures have been proposed. Directional Message Passing Neural Networks (DMPNNs) \cite{gasteiger2020directional} extend the message passing framework by incorporating directional information based on the relative positions of nodes in space. This allows the model to capture the spatial relationships between nodes and learn direction-aware representations.
Equivariant Graph Neural Networks (EGNNs) \cite{satorras2021n} are designed to be equivariant to rotations and translations of the input graph. They achieve this by using equivariant message passing operations and representing node features as high-dimensional vectors that transform according to the group of symmetries.

GNNs have emerged as a powerful and effective framework for analyzing graph-structured data. However, GNNs also face challenges such as the over-smoothing problem, over-squashing issue, and difficulty in capturing long-range dependencies. Despite these limitations, ongoing research aims to address these challenges and further advance the field of GNNs, enabling their application to a wide range of real-world problems involving graph-structured data.

\subsection{Graph Transformer}
Graph Transformers aim to leverage the power of self-attention mechanisms to capture long-term dependencies among nodes while incorporating graph structural information. Existing graph transformer models can be categorized into the following three groups.

\textbf{Designing the architecture of graph Transformer}.  Depending on the relative positions of GNN and Transformer layers, three typical designs have been proposed: (a) building Transformer blocks on top of GNN blocks, e.g., GraphTrans~\cite{wu2021representing}, GraphiT~\cite{mialon2021graphit} and Grover~\cite{rong2020self}, (b) alternately stacking GNN and Transformer blocks, e.g., Mesh Graphormer~\cite{lin2021mesh}, and (c) running GNN and Transformer blocks in parallel and combining their outputs, e.g., Graph-BERT~\cite{zhang2020graph}.

\textbf{Improving positional embeddings with graph structural information. }
 For example, Graph Transformer~\cite{dwivedi2020generalization} proposes to use Laplacian eigenvectors as positional embeddings, which are defined by the eigendecomposition of the graph Laplacian matrix. Other methods, such as Graphormer~\cite{ying2021transformers} and Graph-BERT~\cite{zhang2020graph}, propose to use degree centrality and other heuristic measures as additional positional embeddings.
 
\textbf{Modifying attention mechanisms based on graph priors.} The third group of Graph Transformers aims to modify the attention mechanisms by injecting graph priors. One common approach is to mask the attention matrix based on the graph structure, allowing each node to attend only to its neighbors~\cite{dwivedi2020generalization}. Another approach is to add graph-based bias terms to the attention scores, such as the spatial bias in Graphormer~\cite{ying2021transformers} and the proximity-enhanced multi-head attention in Gophormer~\cite{zhao2021gophormer}.

Graph Transformer models have achieved remarkable success in various domains. However, the optimal way to incorporate graph information into Transformers remains an open question, and the choice of architecture may depend on the specific task and dataset at hand.
\section{Expressive power}
\label{sec:exp_power}
In deep learning theory, the term ``expressive power'' is often used interchangeably with function approximation capability. However, defining the expressive power of Graph Neural Networks (GNNs) in graph learning proves challenging due to the intricate nature of graph-related tasks. Some studies, inspired by deep learning methodologies, explore functions that GNNs can effectively approximate. Alternatively, a conventional approach involves assessing the capacity of GNNs to differentiate non-isomorphic graphs, a fundamental challenge in graph learning. Additionally, certain research endeavors connect the expressive power of GNNs to combinatorial problems or the computation of specific graph properties. These investigations are intimately linked and offer valuable insights into understanding the expressive capabilities of GNNs within the context of graph-based tasks.

In this section, we will elaborate on the theory of the expressive power of GNNs. The hierarchy of the WL algorithm for graph isomorphism problem is the most intuitive measurement and it is the mainstream approach to describe and compare the expressive power of different GNN models. From this point, there are various methods to devise expressive GNNs that are more powerful than 1-WL and we provide their corresponding theoretical results. Finally, we return to the approximation ability that is fundamental for the expressive power of neural networks in deep learning to analyze the universality of GNN.
\subsection{Notations} 
Before reviewing the the theory of expressive power, we introduce some basic notations. $\mathcal{G}=(V,E)$ denotes a graph where $V=\{v_1,v_2,\ldots,v_n\}$ is the node set and $E\subseteq V\times V$ is the edge set. $\mathbf{A}\in \{0,1\}^{N\times N}$ denotes the adjacency matrix and $\tilde{\mathbf{A}} = \mathbf{A} + \mathbf{I}$ denotes the adjacency matrix considering self-loops. The Laplacian matrix of an undirected graph is defined as $\mathbf{L}=\mathbf{D}-\mathbf{A}$ where $\mathbf{D}\in \mathbb{R}^{N\times N}$ is the degree matrix of $\mathbf{A}$ with $\mathbf{D}_{ii}=\sum_{j=1}^N\mathbf{A}_{ij}$. The degree matrix and Laplacian matrix of $\tilde{\mathbf{A}}$ is denoted as $\tilde{\mathbf{D}}$ and $\tilde{\mathbf{L}}=\tilde{\mathbf{D}}-\tilde{\mathbf{A}}$ respectively. $\widehat{\mathbf{A}}=\tilde{\mathbf{D}}^{-\frac{1}{2}}\tilde{\mathbf{A}}\tilde{\mathbf{D}}^{-\frac{1}{2}}$ denotes the normalized $\tilde{\mathbf{A}}$. If available, $\mathbf{X}$ denotes the initial feature matrix of the nodes and $x_v^{(l)}$ denotes the embedding of node $v$ in $l$-th layer. $\mathcal{N}(v)$ denotes the neighbors of $v$. $\{\ldots\}$ denotes the sets while $\{\{\ldots\}\}$ denotes the multi-sets.

\subsection{Graph isomorphism problem and WL algorithm}

\begin{figure}
    \centering
    \includegraphics[width=0.45\textwidth]{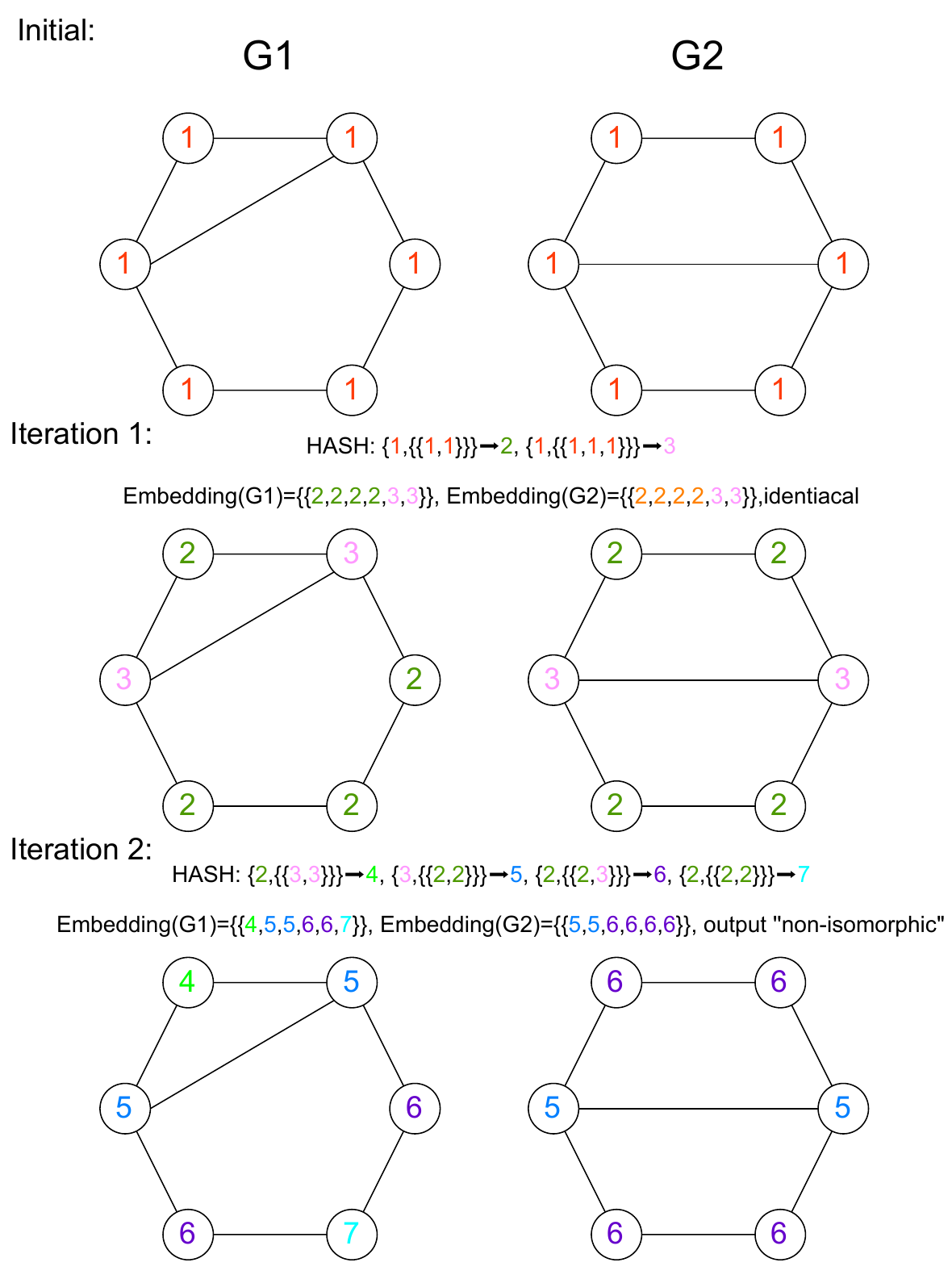}
    \caption{Illustration of 1-WL in distinguishing non-isomorphic graphs within 2 iterations.}
    \vspace{-0.3cm}
    \label{fig:enter-label}
\end{figure}

The graph isomorphism problem involves determining whether two graphs, $\mathcal{G}_1$ and $\mathcal{G}_2$, are structurally identical. Formally, two graphs are considered isomorphic if a bijection exists between their nodes such that edges in $\mathcal{G}_1$ are preserved in $\mathcal{G}_2$.

To deal with the graph isomorphism problem, the Weisfeiler-Lehman (WL) algorithm~\cite{leman1968reduction} is a well-known heuristic algorithm that can be implemented effectively. Its classical form, 1-dimensional Weisfeiler-Lehman (1-WL), also known as color refinement algorithm assigns a label to each node initially and iteratively updates the label via aggregating information from its neighbors. The procedure of the 1-WL algorithm is given in Algorithm \ref{alg:1-WL}, where the HASH function plays the most crucial part as it is required to be injective to guarantee that the different neighborhood information can map to different labels. The Figure~\ref{fig:enter-label} provides an illustration of 1-WL in distinguishing non-isomorphic graphs within 2 iterations. Initially, two non-isomorphic graphs $G_1$ and $G_2$ are given without node features, thus the embedding of each node is set to be identical. In the first iteration, each node pass the embedding of itself together with the multi-set of embedding of the neighbor nodes through an injective HASH function, which obtain the novel embedding representing the degree of the node. Since the two graph $G_1$ and $G_2$ have the same degree distribution, single iteration of 1-WL cannot distinguish them. In the second iteration, the same operation is implemented again but on novel embeddings. This time the two graphs $G_1$ and $G_2$ get different node embedding distributions so the algorithm outputs 'non-isomorphic', which means that two non-isomorphic graphs $G_1$ and $G_2$ are distinguished by the 1-WL in the second iteration.

\begin{algorithm}[t] 
\small
\begin{flushleft}
 \textbf{Input:} A pair of graphs $\mathcal{G}_1=(V,E,X)$ and $\mathcal{G}_2=(U,F,Y)$\\
 \textbf{Output:} cannot determine or non-isomorphic
\end{flushleft}
\begin{algorithmic}[1]
   \STATE $c_v^{(0)}\leftarrow$ HASH$(X_v)$ $ \quad(\forall v\in V)$
   \STATE $d_u^{(0)}\leftarrow$ HASH$(Y_u)$ $ \quad(\forall u\in U)$
   \STATE \textbf{for} $l=1,2,...$(until convergence) \textbf{do}
   \INDSTATE if $\{\{c_v^{(l-1)}|v\in V\}\}\neq\{\{d_u^{(l-1)}|u\in U\}\}$ then return 'non-isomorphic'
   \INDSTATE $c_v^{(l)}\leftarrow$ HASH$(c_v^{(l-1)},\{\{c_w^{(l-1)}|w\in \mathcal{N}_{\mathcal{G}_1}(v)\}\})\quad(\forall v\in V)$
   \INDSTATE $d_u^{(l)}\leftarrow$ HASH$(d_u^{(l-1)},\{\{d_w^{(l-1)}|w\in \mathcal{N}_{\mathcal{G}_2}(u)\}\})\quad(\forall u\in U)$
   \STATE \textbf{end for}
   \STATE return 'cannot determine'
  \end{algorithmic} 
  \caption{1-dimensional WL algorithm(color refinement)}
  \label{alg:1-WL}
\end{algorithm}

The 1-WL algorithm terminates in 
$\mathcal{O}(|U|+|V|)$ iterations and has been shown to effectively distinguish any pair of non-isomorphic random graphs with high probability as the graph size approaches infinity. However, it may struggle to differentiate certain classes of non-isomorphic graphs, like regular graphs of the same order. To address this limitation, more powerful algorithms capable of distinguishing a broader range of non-isomorphic graphs are desired. One such advancement is the $k$-dimensional Weisfeiler-Lehman ($k$-WL) algorithm, which extends the capabilities of the 1-WL algorithm by assigning labels to each $k$-tuple of nodes and the set of k-tuple of nodes is denoted as $V^k$. In the $k$-WL algorithm, the $i$-th neighbor of a $k$-tuple is defined by replacing the $i$-th element in the $k$-tuple with every node in the graph. This approach enhances the expressive power of the algorithm compared to 1-WL, allowing for more robust differentiation between complex graph structures. Additionally, the $k$-dimensional folklore Weisfeiler-Lehman ($k$-FWL) algorithm is another extension that shares similarities with $k$-WL but differs slightly in the aggregation of $k$-tuples. These advancements in the Weisfeiler-Lehman framework offer improved capabilities for distinguishing non-isomorphic graphs and contribute to enhancing the overall performance and versatility of graph isomorphism testing algorithms. The procedure of the $k$-WL is given in Algorithm \ref{alg:k-WL}.
\begin{algorithm}[t] 
\small
\begin{flushleft}
 \textbf{Input:} A pair of graphs $\mathcal{G}_2=(V,E,X)$ and $\mathcal{G}_2=(U,F,Y)$\\
 \textbf{Output:} cannot determine or non-isomorphic
\end{flushleft}
\begin{algorithmic}[1]
   \STATE $c_v^{(0)}\leftarrow$ HASH$(X_v) \quad(\forall v\in V^k)$
   \STATE $d_u^{(0)}\leftarrow$ HASH$(Y_u) \quad(\forall u\in U^k)$
   \STATE \textbf{for} $l=1,2,...$(until convergence) \textbf{do}
   \INDSTATE if $\{\{c_v^{(l-1)}|v\in V^k\}\}\neq\{\{d_u^{(l-1)}|u\in U^k\}\}$ then return 'non-isomorphic'
   \INDSTATE $c_{v,i}^{(l)}\leftarrow\{\{c_w^{(l-1)}|w\in \mathcal{N}_{\mathcal{G}_1,i}(v)\}\}\quad(\forall v\in V^k,i\in [k])$ \tcp{ $\mathcal{N}_{\mathcal{G},i}(v=(v_1,v_2,\ldots,v_k))=\{(v_1,v_2,\ldots,v_{i-1},w,v_{i+1},\ldots,v_k)|w\in V\}$.}
   \INDSTATE $c_v^{(l)}\leftarrow$ HASH$(c_v^{(l-1)},c_{v,1}^{(l)},c_{v,2}^{(l)},\ldots,c_{v,k}^{(l)})\quad(\forall v\in V^k)$
   \INDSTATE $d_{u,i}^{(l)}\leftarrow\{\{d_w^{(l-1)}|w\in \mathcal{N}_{\mathcal{G}_2,i}(u)\}\}\quad(\forall u\in U^k,i\in [k])$
   \INDSTATE $d_u^{(l)}\leftarrow$ HASH$(d_u^{(l-1)},d_{u,1}^{(l)},d_{u,2}^{(l)},\ldots,d_{u,k}^{(l)})\quad(\forall u\in U^k)$
   \STATE \textbf{end for}
   \STATE return 'cannot determine'
  \end{algorithmic} 
  \caption{$k$-dimensional WL algorithm}
  \label{alg:k-WL}
\end{algorithm}

Even though the $k$-WL is more powerful than 1-WL and actually increasing $k$ can obtain more powerful algorithm in distinguishing non-isomorphic graphs, the algorithm has its limitation since there always exists a pair of non-isomorphic graphs for each $k$ such that the $k$-WL algorithm outputs 'cannot determine'.

At the end of the subsection, we provide some useful results about the expressive power of WL algorithm and its variants~\cite{grohe2015pebble} which will be utilized later:
\begin{itemize}
    \item 1-WL and  2-WL have equal expressive power.
    \item $k$-FWL is equivalent to ($k$+1)-WL and thus has equal expressive power.
    \item For $k\geq 2$, ($k$+1)-WL is strictly more powerful than $k$-WL which means that there exists a pair of non-isomorphic graphs that ($k$+1)-WL can distinguish but $k$-WL can not. This implies that the WL algorithm naturally forms a hierarchy.
\end{itemize}

\subsection{Connect GNN with 1-WL}
It is noticed that the role of neighbor aggregation and update in GNN is analogous to that of the hash function operated on one node and its neighbor in 1-WL. Inspired by this close relationship, Xu et al.~\cite{xu2018powerful} first study the expressive power of GNNs with a theoretical framework by viewing the message passing on a node as a rooted subtree and representing the set of its neighbor feature by a multiset. In this way, the aggregation function of GNN can be seen as functions defined on multisets. The authors compare the expressive power between MPNN and 1-WL in distinguishing non-isomorphic graphs and conclude MPNN is at most as powerful as the 1-WL, which is formally given as follows:
\begin{theorem}[Expressive power of MPNN]\label{th:mpnn}
Let $\mathcal{G}_1$ and $\mathcal{G}_2$ be any two non-isomorphic graphs. If a message passing GNN maps $\mathcal{G}_1$ and $\mathcal{G}_2$ to different embeddings, the 1-WL also decides $\mathcal{G}_1$ and $\mathcal{G}_2$ to be non-isomorphic. 
\end{theorem}

Since the aggregation and readout function in GNN are not necessarily injective, the Theorem~\ref{th:mpnn} holds. Further, they prove that if the neighbor aggregation function and graph readout function are injective, the GNN is exactly as powerful as 1-WL. Based on the condition and theory of multisets, they devise a novel GNN architecture named Graph Isomorphism Network (GIN) that is exactly as powerful as the 1-WL. 

\textbf{Graph Isomorphism networks(GINs)}~\cite{xu2018powerful}:
\begin{equation}
    x_v^{(k)}=MLP^{(k)}((1+\epsilon^{(k)})x_v^{(k-1)}+\sum_{u\in \mathcal{N}(v)}x_u^{(k-1)}),
    \label{eq:gin}
\end{equation}
where $\epsilon$ is a learnable scalar value. It is noted that the sum aggregators in Equation~\ref{eq:gin} over MLP are universal functions over multiset and thus can represent injective functions and be adopted as the injective graph readout function in graph classification as well.

Concurrently, Morris et al.~\cite{morris2019weisfeiler} also prove that MPNNs are at most as powerful as the 1-WL. In addition, they state that MPNNs can have the same representation power of 1-WL with proper choosing of the parameter matrices.
\subsection{GNNs beyond 1-WL}
Although previous works have built GNNs that are as powerful as 1-WL, they have weaknesses due to the limited expressive power of 1-WL. For example, they cannot distinguish some pairs of graphs with any parameter matrix unless they have identical labels. More severely, they fail to count simple graph substructures such as triangles~\cite{chen2020can} which is of great importance in computational chemistry~\cite{duvenaud2015convolutional} and social network analysis~\cite{jiang2010finding}. Therefore, many works try to devise GNNs beyond the expressive power of 1-WL.

\subsubsection{High-order GNNs} 
\textbf{$k$-WL based.} One straightforward way to build GNNs beyond 1-WL is resorting to $k$-WL. Morris et al.~\cite{morris2019weisfeiler} propose $k$-GNNs based on set $k$-WL algorithm which is a variant of $k$-WL. Literally, they consider the subgraphs of $k$ nodes instead of $k$-tuple of nodes to make the algorithm more efficient and take the graph structure information into account. To be specific, the set containing all $k$-sets of $V$ is denoted as $[V]_k = \{S\subseteq V| |S| =k\}$ which are the subgraphs of $k$ nodes. In addition, the neighbor of a $k$-set $S$ is defined as the $k$-set with exactly one different node i.e. $\mathcal{N}_{V,k}(S)=\{J \in [V]_k | |J\cap S|=k-1\}$. Although the set $k$-WL is weaker than $k$-WL, it has its own advantage that is more powerful than 1-WL and more scalable than $k$-WL.

In $k$-GNNs, the embedding of the subgraph $S$ of $k$ nodes in layer $t$ is denoted by $x_k^{(t)}(S)$ and the initial feature assigned to each subgraph induced by $S$ represents the isomorphic type of the corresponding subgraph. Then the embeddings of $k$-GNNs can be updated by a message passing scheme according to the Equation (\ref{kgnnup}).
\begin{equation}\label{kgnnup}
    x_k^{(t)}(S) = \sigma(\mathbf{W}_1^{(t)} x_k^{(t-1)}(S)+\sum_{U\in \mathcal{N}_{V,k}(S)}\mathbf{W}_2^{(t)}x_k^{(t-1)}(U)).
\end{equation}

Since the set $k$-WL is more powerful than 1-WL, the $k$-GNN is more powerful than MPNNs and has proven to be as powerful as set $k$-WL with suitable initialization of parameter matrices. The expressive power of the $k$-GNN is characterized by
Theorem \ref{thkgnn} given as follows.

\begin{theorem}[Expressive power of $k$-GNN]\label{thkgnn}
Let $\mathcal{G}_1$ and $\mathcal{G}_2$ be any two non-isomorphic graphs. If a k-GNN maps $\mathcal{G}_1$ and $\mathcal{G}_2$ to different embeddings, the k-set WL also decides $\mathcal{G}_1$ and $\mathcal{G}_2$ to be non-isomorphic. 
\end{theorem}

\textbf{Invariant and equivariant layer based.} 
Graph Neural Networks constructed with high-order tensors offer a novel strategy to address the constraints associated with the 1-WL algorithm. Demanding the representation of a graph remain unchanged under permutations of nodes (invariance) and ensuring node representations reflect consistent transformations corresponding to node reordering (equivariance) respectively, invariance and equivariance stand as pivotal tenets in invariant graph learning.

We use $S_n$ to denote the symmetry group acting on $[n]=\{1,2,\ldots,n\}$ and $\mathbb{R}^{n^k}$ to denote the set of $k$-order tensors. For a tensor $X\in \mathbb{R}^{n^k}$ and a permutation $\sigma \in S_n$, we define the permutation on the tensor as $(\sigma \cdot X)_{\sigma(i_1),\sigma(i_2),\ldots,\sigma(i_k)}=X_{i_1,i_2,\ldots,i_k}$. Then the invariant and equivariant functions can be defined formally as follows.

\begin{definition}[Invariant function] A function $f:\mathbb{R}^{n^k}\rightarrow \mathbb{R}$ is said to be invariant if $f(\sigma\cdot X)=f(X)$ for every permutation $\sigma \in S_n$ and every $X \in \mathbb{R}^{n^k}$.
\end{definition}

\begin{definition}[Equivariant function] A function $f:\mathbb{R}^{n^k}\rightarrow \mathbb{R}^{n^l}$ is said to be equivariant if $f(\sigma \cdot X)=\sigma \cdot f(X)$ for every permutation $\sigma \in S_n$ and every $X \in \mathbb{R}^{n^k}$.\footnote{if $l\neq k$, $\sigma$ also needs to be mapped to a group representation in the target space}
\end{definition}

Note that, in graph learning, the $X\in \mathbb{R}^{n^k}$ is the tensor representation of the graph, and each $k$-tuple $(i_1,i_2,\ldots,i_k)$ can be seen as a hyperedge in the graph. For example, for $k=2$, the adjacency matrix is a 2-order tensor representation of the graph and $X_{ij}$ indicates the existence of edge $(i,j)$. When attaching a feature vector of dimension $d$ to each hyperedge, the tensor is represented by $X\in \mathbb{R}^{n^k\times d}$. Since the permutation is only defined on node indices, i.e. , $(\sigma \cdot X)_{\sigma(i_1),\sigma(i_2),\ldots,\sigma(i_k),i_{k+1})}=X_{i_1,i_2,\ldots,i_k,i_{k+1}}$, the invariant function $f:\mathbb{R}^{n^k\times d}\rightarrow \mathbb{R}$ and equivariant function $f:\mathbb{R}^{n^k\times d}\rightarrow \mathbb{R}^{n^l \times d}$ follow similar modification in definition.

Maron et al.~\cite{maron2018invariant} provide a full characterization of all linear invariant and equivariant layers acting on a $k$-order tensor for the first time by solving the fixed-point equation of the permutation matrix group. Specifically, for layers devoid of bias and features, the dimension of the linear invariant and equivariant layer space is precisely given as follows:
\begin{theorem}[Dimension of linear invariant and equivariant layers]\label{space} The space of invariant linear layer $L:\mathbb{R}^{n^k}\rightarrow \mathbb{R}$ and equivariant linear layer $L:\mathbb{R}^{n^k}\rightarrow \mathbb{R}^{n^l}$ are of dimension $b(k)$ and $b(k+l))$ respectively, where $b(k)$ is the $k$-th Bell number that represents the number of ways a set of n elements can be partitioned into non-empty subsets.
\end{theorem}

From Theorem~\ref{space}, it is surprising to find that the dimension of the space is independent of the size of the graph which enables us to apply the same GNN with a given order of linear invariant and equivariant layers to graphs of any size. The Theorem~\ref{space} can be further generalized to the layers with bias and features or multi-node sets and derive similar results. For more detailed information, readers can refer to the original paper.

With the formula of all linear invariant and equivariant layers, Maron et al.~\cite{maron2018invariant} prove that the GNN built by the layers can approximate any message passing network to an arbitrary precision on a compact set, which implies that the proposed model is at least as powerful as MPNN. Further, they proposed a new GNN architecture called $k$-order invariant graph network $F$:
\begin{equation}\label{kinvariantgnn}
    F=m\circ L_I \circ L_d \circ \phi \circ \ldots \circ \phi \circ L_1,
\end{equation}
where $L_i:\mathbb{R}^{n^{k_i}\times a_i}\rightarrow \mathbb{R}^{n^{k_{i+1}}\times a_{i+1}},\max_{i\in [d+1]}k_i=k$ are equivariant linear layers, $a_i$ denotes the dimension of feature in the $l$-th layer, $\phi$ is an activation function, $
L_I:\mathbb{R}^{n^{k_{d+1}}\times a_{d+1}}\rightarrow \mathbb{R}^{a_{d+2}}$ is an invariant linear layer and $m:\mathbb{R}^{a_{d+2}}\rightarrow \mathbb{R}^{a_{d+3}}$ is a multilayer perception. The $k$-order GNN $F$ is provably able to encode the multisets computed in the $k$-WL with a suitable weight matrix and thus can implement the $k$-WL, which leads to the Theorem \ref{kordergnn}.

\begin{theorem}[Expressive power $k$-order GNN]\label{kordergnn} Given two graphs $\mathcal{G}_1$ and $\mathcal{G}_2$. If the two graphs can be distinguished by the $k$-WL, there exists a $k$-order network $F$ such that $F(\mathcal{G}_1)\neq F(\mathcal{G}_2)$. On the other direction for every two isomorphic graphs $G_1$ and $G_2$ and a $k$-order network, we have $F(\mathcal{G}_1)=F(\mathcal{G}_2)$.
\end{theorem} 

Theorem~\ref{kordergnn} indicates that the $k$-order GNN is at least as powerful as the $k$-WL in terms of distinguishing non-isomorphic graphs. However, the $k$-order GNN is impractical for $k\geq 3$ because of the $\mathcal{O}(n^k)$ memory cost. Therefore, the authors propose a simple GNN model based on 2-FWL that is as powerful as 3-WL while only utilizing tensors of order 2. The proposed model replaces the equivariant linear layers and activation functions with specific blocks and the architecture is given as follows.
\begin{equation}\label{fgnn}
    F=m\circ L_I \circ B_d \circ B_{d-1} \circ \ldots \circ B_1
\end{equation}

In each block $B_i$, the authors 
 apply three MLPs that implement matrix multiplication to match the feature then concatenate the embedding to obtain the output tensor. They prove that the matrix multiplication can implement the aggregation used in 2-FWL to boost the expressive power. The technique can be further generalized to $k$-order GNNs to make them as powerful as ($k$+1)-WL. The expressive power of the proposed model named PPGN is presented in Theorem~\ref{thfgnn}.
\begin{theorem}[Expressive power of PPGN]\label{thfgnn}Given two graphs $\mathcal{G}_1$ and 
 $\mathcal{G}_2$. If the two graphs can be distinguished by the 2-FWL (3-WL), there exists a GNN $F$ defined by Equation~\ref{fgnn} such that $F(\mathcal{G}_1)\neq F(\mathcal{G}_2)$. On the other direction for every two isomorphic graphs $\mathcal{G}_1$ and $\mathcal{G}_2$ and the model $F$ defined by Equation~\ref{fgnn}, we have $F(\mathcal{G}_1)=F(\mathcal{G}_2)$.
\end{theorem} 

Besides, it is noteworthy that some works devise powerful GNNs with polynomial layers that are also able to preserve the invariant and equivariant property~\cite{maron2019provably,chen2019equivalence,azizian2020expressive}. Recently, Puny et al.~\cite{puny2023equivariant} formalize the equivariant graph polynomial that is a matrix polynomial map equivariant to node permutation. The authors further propose the polynomial hierarchy that alleviate some problems of WL hierarchy and provide a full characterization of all the graph polynomials. Equipping with the polynomial features, the PPGN~\cite{maron2019provably} can be strictly more powerful than 3-WL while only costing $O(n^2)$ memory.

\textbf{Local and sparsity-aware high-order GNN.} A key limitation of prior approaches of high order GNNs is the heavy computation and memory cost. The methods based on $k$-WL consider all the tuples or subgraphs of $k$ nodes while linear invariant or equivariant layers are defined in the tensor of $k$-order, which all require $O(n^k)$ memory. Besides, the model defined by Equation~\ref{fgnn} faces similar challenges due to the multiplication operation of dense matrices. 

Hence, certain studies propose adaptations of the $k$-Weisfeiler-Lehman ($k$-WL) algorithm that focus on a particular category of $k$-node objects and establish neighborhood relationships locally, aiming to achieve a balance between expressive power and scalability. For instance,  Morris et al.~\cite{morris2020weisfeiler} introduce $\delta$-$k$-LWL and corresponding $\delta$-$k$-LGNN that consider the local and global neighbors. They~\cite{morris2022speqnets} further propose ($k,s$)-LWL and corresponding $(k,s$)-SpeqNet that only act on $k$-tuples which induces subgraphs of at most $s$ connected components to reduce the computation cost greatly. The $(k,c)(\leq)$-SETWL and $(k,c)(\leq)$-SETGNN proposed by Zhao et al.~\cite{zhao2022practical} share a similar idea but they utilize $k$-sets instead of $k$-tuples. Besides, Wang et al.~\cite{wang2022mathscr} introduce $\mathscr{N}(t,d)$-WL and devise corresponding GNN architecture G3N that aggregates the induced subgraphs of $k$ nodes within the $d$-hop neighborhood of a node to fit real-world tasks better. These methods provide strong theoretical foundations in establishing a unique hierarchy for distinguishing non-isomorphic graphs and exhibit efficient applicability to real-world tasks. 

\subsubsection{Graph property based GNNs}
\textbf{Substructure based GNN.} 
In addition to the challenges in distinguishing non-isomorphic graphs, GNNs also encounter obstacles in quantifying simple substructures like triangles and cliques~\cite{chen2020can}, which is of great importance in various real applications such as drug discovery~\cite{duvenaud2015convolutional} and social network studies~\cite{jiang2010finding}. Therefore, the capability to detect and count substructures serves as an intuitive metric to evaluate the expressive power of GNNs. 

Chen et al.~\cite{chen2020can} initiate the exploration by providing a theoretical framework for studying the expressive power of GNNs via substructure counting. Specifically, they define two types of counting on attribute graphs: containment-count and matching-count, representing the number of subgraphs and induced subgraphs isomorphic to a specified substructure respectively. 
\begin{definition}[Containment-count and matching-count] Let $\mathcal{G}^P$ be a graph that we refer to as a pattern or substructure. We define $\mathcal{C}(\mathcal{G},\mathcal{G}^P)$, called the containment-count of $\mathcal{G}^P$ in $\mathcal{G}$, to be the number of subgraphs of $\mathcal{G}$ that are isomorphic to $\mathcal{G}^P$. We define $\mathcal{M}(\mathcal{G},\mathcal{G}^P)$, called the matching-count of $\mathcal{G}^P$ in $\mathcal{G}$, to be a number of induced subgraphs of $\mathcal{G}$ that are isomorphic to $\mathcal{G}^P$.
\end{definition}

Since the induced graphs belong to graphs, $\mathcal{M}(\mathcal{G},\mathcal{G}^P)\leq \mathcal{C}(\mathcal{G},\mathcal{G}^P)$ always holds. With this framework, they analyze the previous GNN architectures and WL algorithm concerning the two substructure counting criteria. The derived results are given below.
\begin{itemize}
    \item 1-WL, 2-WL and 2-IGN cannot perform matching-count of any connected substructures with 3 or more nodes. However, they can perform containment-count of star-shaped substructures.
    \item $k$-WL and $k$-IGN is able to perform both matching-count and containment-count of patterns of $k$ nodes. Besides, running $T$ iterations of $k$-WL cannot perform matching-count of any path substructure of $(k+1)2^T$ or more nodes.
\end{itemize}

Although more general results for $k$-WL are expected and the work does not devise a novel GNN architecture equipped with substructure counting, it provides a solid foundation to measure the expressive power of GNN by substructure counting. Indeed, the high-order GNNs discussed earlier can enumerate certain substructures~\cite{arvind2020weisfeiler,furer2017combinatorial}; however, due to challenges posed by large $k$ values and the incompetency to address more intricate substructures, they turn out to be impractical.

We then focus on introducing GNNs that leverage information garnered from substructure counting to improve their expressive capabilities. For instance, Bouritsas et al.~\cite{bouritsas2022improving} integrate substructure counting into node and edge features, deriving structural attributes by tallying specific substructures. Their Graph Substructure Networks (GSN) exhibit greater expressiveness than 1-WL, effectively distinguishing non-isomorphic graphs beyond the capabilities of 3-WL. Furthermore, they establish sufficient conditions for universality. Barceló et al.~\cite{barcelo2021graph} extend the work of Bouritsas et al.~\cite{bouritsas2022improving} by performing homomorphism counting of substructures. Horn et al.~\cite{horn2021topological} employ graph filtration to capture the emergence and disappearance of specific substructures. Toenshoff et al.~\cite{toenshoff2021graph} implement random walks to detect small substructures. Bodnar et al. investigate substructure counting on simplex\cite{bodnar2021weisfeiler} and regular cell complexes~\cite{bodnar2021weisfeilercw} with lifting transformation. However, it is noteworthy that many of the aforementioned studies focus on specific types of substructures and the selection of substructures is often manual and heuristic, which presents challenges in adapting substructure-based GNNs to real-world applications.

Besides, we highlight two GNN architectures based on subgraphs in advance for their superior performance in counting substructures. Huang et al.~\cite{huang2022boosting} propose a GNN architecture where each node-based subgraph is augmented with a pair of node identifiers. This enables the GNN to count all cycles of length up to 6. More Recently, Tahmaesebi et al.~\cite{tahmasebi2023power} introduce the Recursive Neighborhood Pooling Graph Neural Network (RNP-GNN), which performs recursive pooling on the node-based subgraphs for each node using node marking and neighborhood intersection technique. The authors provide theoretical proof that for any set of substructures, there exists an RNP-GNN that can count them. Recently, Zhou et al.~\cite{zhou2024distance} propose Distance-Restricted FWL GNNs. By restricting the node pairs of 2-FWL to be only those with distance less than 2, their GNN can provably count up to 6 cycles with the best known complexity.

\textbf{Distance based GNN.} It is observed that WL algorithm neglects the distance information for its neighborhood aggregation scheme, thus some works combine WL algorithm or GNN with distance information to enhance the expressive power. Zhang and Chen~\cite{zhang2018link} first leverage the shortest path distances between target nodes and other nodes to enhance the link prediction performance of GNNs in their SEAL algorithm. Li et al.~\cite{li2020distance} generalize it into distance encoding (DE) defined by random walks to learn structural representation and overcome the limitation of 1-WL. Specifically, they utilize the shortest path distance and generalized PageRank Scores~\cite{li2019optimizing} as the measurement of DE, which are further served as extra features or controllers of message aggregation to devise powerful GNN architecture named DE-GNN and DEA-GNN. 

In addition, the $k$-hop MPNN~\cite{nikolentzos2020k} that aggregates the embedding within $k$-hop neighborhood of each node simultaneously can be viewed as another type of GNN using the distance information. Feng et al.~\cite{feng2022powerful} first analyze the expressive power of the $k$-hop MPNN and derive the following theorem.
\begin{theorem}[Expressive power of $k$-hop MPNN] A $k$-hop MPNN with suitable parameters is strictly more powerful than 1-hop MPNN when $k>$1 while the expressive power of a $k$-hop MPNN is bounded by 3-WL.
\end{theorem}

Further, they improve the expressive power of $k$-hop MPNN by equipping the message passing with peripheral subgraph information. The proposed KP-GNN is proven to be capable of distinguishing almost all regular graphs with a proper $k$. 

Recently, Zhang and Luo~\cite{zhang2023rethinking} introduced a novel class of expressive power metrics via graph biconnectivity and show that most existing GNN architectures fail to solve biconnectivity problem. Therefore, they propose a principled and efficient algorithm called the Generalized Distance Weisfeiler-Lehman (GD-WL) to solve the problem. The core step in the algorithm is given as follows.
\begin{equation}
    c_v^{(l)} \leftarrow \text{HASH}({{(d_{uv}^{(l-1)}, c_u^{(l-1)}):u\in V}}),
\end{equation}
where $d_{uv}$ is an arbitrary distance metric. They further propose two appropriate distance metrics to enable the GD-WL to solve all biconnectivity problems. The main results are shown in Theorem ~\ref{biconnect}.

\begin{theorem}[Expressive power of different distance metrics for solving biconnectivity problem]\label{biconnect} SPD (shortest path distance)-WL is fully expressive for edge-biconnectivity and RD (resistance distance)-WL is fully expressive for vertex-biconnectivity. When using both SPD and RD, the obtained Generalized Distance WL Algorithm(GD-WL) is fully expressive for both edge-biconnectivity and vertex-biconnectivity.
\end{theorem}

Furthermore, the authors compare the SPD-WL and RD-WL to existing WL hierarchy and prove that the  expressive power of SPD-WL and RD-WL is bounded by 2-FWL(3-WL), which also indicates that 2-FWL(3-WL) is fully expressive for both edge-biconnectivity and vertex-biconnectivity.

\textbf{Graph spectral based GNN.} There are some works incorporating the spectral information into GNNs and the approach is proven to easily break the limitation of 1-WL. Balcilar~\cite{balcilar2021breaking} analyze the expressive power of GNN from the a spectral perspective and prove that most MPNNs act as low-pass filters which limit their expressive power. To break the limitation and further consider expressive power in terms of other perspectives such as graph isomorphism test and substructure counting, they resort to matrix language (MATLANG) proposed by Geert~\cite{geerts2021expressive} and design graph convolution supports in spectral domain. Besides, Feldman et al.~\cite{feldman2022weisfeiler} utilize spectral feature based on graph Laplacian in an additional pre-coloring phase to improve the expressive power of GNN. Wang and Zhang~\cite{wang2022powerful2} prove that the expressive power of a wide range of spectral GNNs based on $k$-degree polynomial filters is bounded by $k+1$ iterations of WL. Besides, it is worth noting that the spectral information can also be utilized to boost the expressive power of graph transformers~\cite{dwivedi2021graph,kreuzer2021rethinking,kim2022pure} that we will discuss later.

\subsubsection{Subgraph GNNs} 
Motivated by the observation that non-isomorphic graphs always have non-isomorphic subgraphs, the subgraph GNNs have been popular recently for its solid theoretical guarantee as well as flexible design. Typically, the subgraph GNNs can be categorized into node-based and edge-based. The node-based subgraph GNNs are more common in practice while edge-based subgraph GNNs compute the representation of edge-node pairs additionally. Roughly, they implement three steps: subgraph extraction by a specific policy, message passing to obtain individual representation and graph pooling. 

Following the procedure, existing works propose subgraph GNNs with a variety of architectures. Cotta et al.~\cite{cotta2021reconstruction} adopt node removal to generate subgraphs based on graph reconstruction conjecture. Papp and Wattenhofer, ~\cite{papp2022theoretical} prove that node-marking is a more expressive approach than intuitive node removal. Bevilacqua et al.~\cite{bevilacqua2021equivariant} represent each graph as a set of subgraphs and study four simple but effective subgraph selection policies: node-deleted subgraphs, edge-deleted subgraphs, and two corresponding variants of ego-networks. Wijesinghe and Wang~\cite{Wijesinghe2022new} incorporate the local structure captured by overlap subgraphs into a message passing scheme to obtain GNN architecture that is more powerful than 1-WL. You et al.~\cite{you2021identity} add identity information to the center node of each subgraph to break the symmetry while Huang et al.~\cite{huang2022boosting} implement pairs of node identifiers assigned to the center node and one of its neighborhoods. Zhang et al.~\cite{zhang2021nested} perform message passing on rooted subgraphs around each node instead of the rooted subtree. Similar to Zhang's work, Zhao et al.~\cite{zhao2021stars} utilize a base GNN as a kernel to encode the star subgraph of each node to generate multiple subgraph-node embeddings. Thiede et al.~\cite{thiede2021autobahn} consider an automorphism group of subgraphs to construct expressive GNNs.

Besides the various methods to design powerful subgraph GNNs, there are some works that focus on analyzing the expressive power of subgraph GNNs. Frasca et al.~\cite{frasca2022understanding} provides an extensive analysis of the expressive power of node-based subgraph GNNs from the perspective of symmetry. They observe that the node-based policies define a bijection between nodes and subgraphs thus the expressive power of node-based subgraphs GNNs can be characterized by one single permutation group acting jointly on nodes and subgraphs while previous works often consider two permutation groups defined on nodes and subgraphs separately. Besides, The symmetry structure described by the new permutation group is highly consistent with that of IGN. With this observation, they bound the expressive power of node-based subgraphs GNNs as follows.

\begin{theorem}[Expressive power of 3-IGN] The 3-IGN can implement the node-based policies of subgraph GNNs thus the expressive power of node-based subgraph GNNs is bounded by 3-IGN which is proved to be as powerful as 2-FWL(3-WL).
\end{theorem}

Qian et al.~\cite{qian2022ordered} introduce a unified theoretical framework for studying various designs of subgraph GNNs via a new variant of 1-WL called $k$-ordered subgraph WL ($k$-OSWL). The implementation of $k$-OSWL is similar to a procedure that performs 1-WL on each of the $k$-ordered subgraphs and then aggregates all the $k$-ordered subgraphs to compute the embedding for each node. They analyze the expressive power of the proposed $k$-OSWL and compare it with the original $k$-WL.
\begin{theorem}[Expressive power of $k$-OSAN] The expressive power of $k$-OSANs is bounded by ($k$+1)-WL but it is incomparable to $k$-WL. Besides, ($k$+1)-OSANs is strictly more powerful than $k$-OSANs, which forms a hierarchy. 
\end{theorem}

Although the expressive power of $k$-OSANs is incomparable to WL algorithm with the same order, the hierarchy of $k$-OSANs suggests that increasing the size of the subgraphs can boost the expressive power. Zhou et al.~\cite{zhou2023relational} further generalize $k$-OSWL's running 1-WL on k-ordered subgraph to $k,l$-WL, which runs $k$-WL on $l$-ordered subgraph. They characterize the $k,l$-WL hierarchy by comparing it with $k$-WL, and prove that $k,l$-WL is less expressive than $k+l$-WL.

Zhang et al.~\cite{zhang2023complete} provide a systematic characterization of the expressive power of node-based subgraph GNNs via a new version of WL hierarchy called Subgraph WL (SWL). Specifically, they categorize any node-based subgraph GNN into one of six equivalence classes according to different strategies used in subgraph generation, equivariant message passing, and final pooling. Among these six equivalence classes, they prove that the node marking SSWL using both the local aggregation and vertex-subgraph pooling achieves the maximal expressive power. To relate the proposed SWL to the existing WL hierarchy and provide a more precise hierarchy, they introduce a localized version of FWL algorithms and utilize the pebbling game framework~\cite{ehrenfeucht1961application,fraisse1955quelques} to compare the expressive power of different algorithms. With this framework, they make a strict comparison between different equivalence classes in the SWL and derive a tight expressive power upper bound to the localized FWL and FWL. It is noted that most of the above-mentioned methods and theories only consider the node-based subgraph GNN while the counterparts of the other type of subgraph GNNs, i.e. edge-based subgraph GNNs are rarely explored. Therefore, characterizing the expressive power of edge-based subgraph GNNs and further revealing the relation between the two types of subgraph GNNs may be a possible direction for future study.

\subsubsection{Non-equivariant GNNs} Resorting to some non-equivariant operations can directly break the symmetry of MPNN, thus enhancing the expressive power of GNNs to go beyond 1-WL. For instance, the relational pooling~\cite{murphy2019relational} inspired by joint exchangeability~\cite{aldous1981representations} is inherently permutation-invariant for taking an average of all permutations on a graph. Formally, the relational pooling obtains the embedding of the graph $\mathcal{G}$ with an arbitrary function $f$ as follows.
\begin{equation}
    \overline{\overline{f}}(\mathbf{A},\mathbf{X})=\frac{1}{n!}\sum_{\sigma \in S_n} f(\sigma\cdot \mathbf{A},\sigma\cdot \mathbf{X}),
\end{equation}
where $\sigma$ is the permutation defined on the symmetric group $S_n$.
To improve the expressive power of GNN with the relational pooling, the authors attach each node a permutation-sensitive identifier thus making the method non-equivariant, which can formulated as concatenating a one-hot encoding to the feature. The derived novel GNN architecture called RP-GNN is defined as follows. 
\begin{equation}\label{RPGNN}
    \overline{\overline{f}}(\mathbf{A},\mathbf{X})=\frac{1}{n!}\sum_{\sigma \in S_n} f(\mathbf{A},[\mathbf{X},\sigma\cdot \mathbf{I}_n]),
\end{equation}
where $\mathbf{I}_n \in \mathbb{R}^{n \times n}$ is the identity matrix and we omit the permutation acting on the graph since the GNN is permutation-invariant.

Based on the the Equation~\ref{RPGNN}, the authors further prove that the RP-GNN is strictly more powerful than the original GNN in terms of distinguishing non-isomorphic graphs, which provides a practical tool to boost the expressive power GNN. Therefore, equipping the GIN with the relational pooling can easily derive a GNN that is more powerful than 1-WL. In addition, the local relational pooling is able to help GNNs count triangles and 3-stars empirically~\cite{chen2020can}.

Besides the relational pooling, there are some other intuitive non-equivariant techniques to increase the expressive power of GNN. Papp et al.~\cite{papp2021dropgnn} utilize dropout techniques to remove a certain proportion of nodes during the train and test phase. Sato et al.~\cite{sato2021random} and Abbound et al.~\cite{abboud2021surprising} add random features drawn from a standard uniform distribution to the initialization of node features. Sato et al.~\cite{sato2019approximation} introduces port numbering that is widely used in distributed local algorithms to GNN which we will discuss in the next subsection. Those non-equivariant techniques are easy to implement but their performance cannot always be guaranteed since they do not preserve the permutation equivariance property of GNNs.

\subsection{Connect GNN with combinatorial problems} 
Besides the graph isomorphism problem, GNNs have been used to solve some NP-hard combinatorial problems in recent years, including minimum dominating set problem and minimum vertex cover problem~\cite{khalil2017learning,li2018combinatorial}. Since those problems cannot be solved in polynomial time concerning the input size if we assume that $P\neq NP$, the GNNs are merely able to provide sub-optimal solutions with certain approximation ratios. Therefore, it is also feasible to analyze the expressive power of GNNs by the approximation ratio that they can achieve for those combinatorial problems~\cite{sato2019approximation}.

To better reveal the role of GNNs in combinatorial problems, Sato et al.~\cite{sato2019approximation} connect GNNs to distributed local algorithm~\cite{suomela2013survey} that is efficient in solving combinatorial problems and specify two classes of GNNs: multiset-broadcasting GNN (MB-GNN) and set-broadcasting GNN (SB-GNN) as follows. 
\begin{definition}[MB-GNN and SB-GNN]\label{MBGNN_SBGNN}
    \begin{align}
    x_v^{(k)} &= f(\{\{x_u^{(k-1)}|u\in \mathcal{N}(v)\}\}), \quad \text{(MB-GNN)} ,\\
    x_v^{(k)} &= f(\{x_u^{(k-1)}|u\in \mathcal{N}(v)\}), \quad \text{(SB-GNN)}.
\end{align}
\end{definition}

According to Definition~\ref{MBGNN_SBGNN}, the MB-GNN corresponds to the MPNN while SB-GNN is a special class of MB-GNN that restricts the aggregated embeddings to be a set. To break the symmetry of message passing, the authors introduce port numbering that is widely used in distributed algorithms to GNNs that enables the GNN to send different messages to different neighbors, which obtains a new class of GNN named vector-vector consistent GNN($VV_C-GNN$) that is strictly more powerful than previous MB-GNNs. The $VV_C-GNNs$ updates the node feature as follows.
\begin{equation}
    x_v^{(k)} = f({p(u,v),p(v,u),x_u^{(k-1)}|u\in \mathcal{N}(v)}) \quad (VV_C-GNN),
\end{equation}
where $p(v,u)$ is the port number of $v$ that edge $(v,u)$ connects to. Further, the authors propose the most powerful $VV_C-GNN$ called consistent port numbering GNNs(CPNGNNs) that aggregate the features by concatenation. The theorem given as follows demonstrates the performance of CPNGNNs in solving combinatorial problems. 
\begin{theorem}[Approximation ratio of CPNGNN in solving combinatorial problems]\label{CPNGNN} CPNGNNs can achieve at most $\Delta$+1-approximation for the minimum dominating set problem and at most 2-approximation for the minimum vertex cover problem where $\Delta$ is the maximum degree in the input graph.
\end{theorem}

Although the derived approximation ratio is far from optimal, it can be further improved by additional information about the graph. Later, Sato et al.~\cite{sato2021random} proposed a simple but efficient technique to boost the expressive power of GNN by concatenating a random feature sampled i.i.d. form a uniform distribution to the initial feature. Equipped With this slight modification, the authors prove that the GIN can achieve a near-optimal approximation ratio for the minimum dominating set problem and minimum vertex cover problem.

\subsection{Approximation ability of GNN}

Having explored the expressive capabilities of GNNs across diverse graph-related tasks thus far, this subsection discusses the approximation theory—an essential framework for describing expressive power within deep learning ~\cite{cybenko1989approximation,hornik1991approximation}. Specifically, our attention shifts towards the graph functions that GNNs can effectively approximate to by analyzing current approximation results. Additionally, we illustrate the close relationship between the approximation ability and the ability to distinguish non-isomorphic graphs.

Since the graph embedding is always assumed to be invariant to the permutation of nodes, it is natural to ask whether a GNN can approximate any invariant functions to evaluate its expressive power. From this point of view, Maron et al.~\cite{maron2019universality} first analyze the expressive power of $G$-invariant networks formulated in Equation~\ref{kinvariantgnn} that are networks with invariant or equivariant linear with respect to arbitrary subgroup $G$ of a symmetric group $S_n$, in terms of approximating any continuous invariant functions. Formally, the universality of $G$-invariant networks is given as follows.

\begin{theorem}[Universality of $G$-invariant GNN]\label{Ginvariant} Let $f:\mathbb{R}^n\rightarrow\mathbb{R}$ be a continuous $G$-invariant function that satisfies $f(\sigma\cdot x) = f(x)$ for all $x\in \mathbb{R}^n$ and $\sigma\in G\leq S_n$, and $K\subset \mathbb{R}^n$ a compact set. Then there exists a $G$-invariant network that can approximate $f$ to an arbitrary precision.
\end{theorem}

The Theorem~\ref{Ginvariant} indicates that any continuous $G$-invariant function can be approximated by a $G$-invariant neural network to an arbitrary precision. In addition, they show that the upper bound of the tensor order for $G$-invariant polynomial to achieve the universal approximation ability is $\frac{n(n-1)}{2}$. Since the upper bound is unfeasible for the expensive computational cost, they provide a practical lower bound of the order that is $\frac{n-1}{2}$ for universality. 

Alternatively, Keriven and Peyre\cite{keriven2019universal} provide a proof of the same result by retaining to a one-layer but similar architecture of Equation~\ref{kinvariantgnn} and further extend the result to equivariant case for the first time. Unlike Maron's work consider a fixed $n$ when analyzing universality, they prove that the GNNs with a single set of parameters can approximate any continuous invariant function uniformly well as long as the graph size is bounded by a specific parameter. Furthermore, Barcel{\'o} et al.~\cite{barcelo2020expressive} derive the completely uniform results that are independent of graph size through the lens of logical classifiers. Similarly, Grohe~\cite{grohe2023descriptive} provides a precise characterization of the the graph functions that can be computed by a class of polynomial-size bounded depth GNNs via Boolean
circuit complexity. In addition, there are some techniques to achieve universality. For instance, Abbound et al.~\cite{abboud2021surprising} prove that the random initialization feature can help MPNN approximate any invariant functions defined on graphs with high probability, which is the first universality result for the MPNN.

Besides the universal approximation, Loukas~\cite{loukas2019graph} consider the expressive power of MPNN with respect to Turing universality, which refers to the ability to compute any function that is computable by a Turing machine with the same input. Compared to the universal approximation, the Turing universality is strictly stronger and able to solve graph isomorphism problems. To obtain sufficient conditions for the MPNN to achieve the Turing universality, the author proves that the MPNN is equivalent to the LOCAL model in the distributed algorithm which is a well-studied Turing universal model. With the established equivalence, the sufficient conditions for MPNN to achieve Turing universality are given as follows:
\begin{itemize}
    \item The GNN should be sufficiently wide and deep.
    \item The functions applied in each layer should be sufficiently expressive.
    \item The nodes can uniquely identify each other.
\end{itemize}
It is noted that the last condition can partly account for the effectiveness of unique node identifiers and other approaches that break the symmetry adopted in the previous subsection.

Finally, we dip into works that bridge the gap between graph isomorphism testing and function approximation, two primary lenses for evaluating the expressive capabilities of GNNs. Chen et al.~\cite{chen2019equivalence} establishes the theoretical equivalence between these perspectives by introducing GIso-discriminating, a novel concept that extends the discrete graph isomorphism problem to function approximation within a continuous input space. Moreover, they propose a structured framework using sigma-algebra terminology to systematically compare the expressive capacities of various models. Azizian et al.~\cite{azizian2020expressive} also delves into expressive power through these two viewpoints, focusing on three categories of GNNs: MPNN, linear GNN and folklore GNN, encompassing both invariant and equivariant instances. Their analysis not only reaffirms previous universality findings but also identifies the $k$-folklore GNN as the most powerful among the three architectures, capable of approximating any continuous invariant function but less powerful than ($k$+1)-WL.

\begin{table*}[t]
\caption{Summary of expressive power of GNN architectures.}
\label{Table:expressive_power}
\small
\resizebox{2\columnwidth}{!}{
\begin{threeparttable}
\begin{tabular}{llccl}
\hline
\multicolumn{1}{c}{\multirow{2}{*}{Category}} & \multicolumn{1}{c}{\multirow{2}{*}{Work}} & \multicolumn{2}{c}{Technique} & \multicolumn{1}{c}{\multirow{2}{*}{Expressive power$^*$}} \\
\multicolumn{1}{c}{} & \multicolumn{1}{c}{} & Node feature & GNN architecture & \multicolumn{1}{c}{} \\ \hline
\multicolumn{1}{l}{MPNN} & Gilmer et al.~\cite{gilmer2017neural} &  &  & not more powerful than 1-WL \\ \hline
\multirow{7}{*}{High-order GNNs} & Xu et al.~\cite{xu2019how} &  & $\checkmark$ & as powerful as 1-WL \\ \cline{2-5} 
 & Morris et al.~\cite{morris2019weisfeiler} &  & $\checkmark$ & not more powerful than k-WL \\ \cline{2-5} 
 & Maron et al.~\cite{maron2019provably} &  & $\checkmark$ & as powerful as k-WL \\ \cline{2-5} 
 & Chen et al.~\cite{chen2019equivalence} &  & $\checkmark$ & more powerful than 1-WL \\ \cline{2-5} 
 & Morris et al.~\cite{morris2020weisfeiler} &  & $\checkmark$ & not more powerful than k-WL \\ \cline{2-5} 
 & Morris et al.~\cite{morris2022speqnets} &  & $\checkmark$ & not more powerful than k-WL \\ \cline{2-5} 
 & Zhao et al.~\cite{zhao2022practical} &  & $\checkmark$ & not more powerful than k-WL \\ \hline
\multirow{8}{*}{Substructure-based GNNs} & Barcelo et al.~\cite{barcelo2021graph} & $\checkmark$ &  & less powerful than k-WL \\ \cline{2-5} 
 & Bodnar et al.~\cite{bodnar2021weisfeilercw} &  & $\checkmark$ & more powerful than 1-WL \\ \cline{2-5} 
 & Bodnar et al.~\cite{bodnar2021weisfeiler} &  & $\checkmark$ & more powerful than 1-WL \\ \cline{2-5} 
 & Thiede et al.~\cite{thiede2021autobahn} &  & $\checkmark$ & not more powerful than 1-WL \\ \cline{2-5} 
 & Horn et al.~\cite{horn2021topological} & $\checkmark$ &  & more powerful than 1-WL \\ \cline{2-5} 
 & Toenshoff et al.~\cite{toenshoff2021graph} & $\checkmark$ & $\checkmark$ & incomparable to 1-WL \\ \cline{2-5} 
 & Bouritsas et al.~\cite{bouritsas2022improving} & $\checkmark$ &  & more powerful than 1-WL \\ \cline{2-5} 
 & Choi et al.~\cite{choi2022cycle} & $\checkmark$ &  & incomparable to WL hierarchy \\ \hline
\multirow{5}{*}{Distance-based GNNs} & Li et al.~\cite{li2020distance} & $\checkmark$ & $\checkmark$ & more powerful than 1-WL \\ \cline{2-5} 
 & Zhang et al.~\cite{zhang2023rethinking} & $\checkmark$ & $\checkmark$ & as powerful as 3-WL \\ \cline{2-5} 
 & Nikolentzos et al.~\cite{nikolentzos2020k} &  & $\checkmark$ & not more powerful than 3-WL \\ \cline{2-5} 
 & Feng et al.~\cite{feng2022powerful} &  & $\checkmark$ & more powerful than 1-WL \\ \cline{2-5} 
 & Wang et al.~\cite{wang2022mathscr} &  & $\checkmark$ & incomparable to k-WL \\ \hline
\multirow{11}{*}{Subgraph GNNs} & Bevilacqua et al.~\cite{bevilacqua2021equivariant} &  & $\checkmark$ & more powerful than 1-WL \\ \cline{2-5} 
 & Papp et al.~\cite{papp2021dropgnn} &  & $\checkmark$ & more powerful than 1-WL \\ \cline{2-5} 
 & Cotta et al.~\cite{cotta2021reconstruction} &  & $\checkmark$ & more powerful than 1-WL \\ \cline{2-5} 
 & Zhao et al.~\cite{zhao2021stars} &  & $\checkmark$ & more powerful than 1-WL \\ \cline{2-5} 
 & Zhang \& Li~\cite{zhang2021nested} &  & $\checkmark$ & more powerful than 1-WL \\ \cline{2-5} 
 & You et al.~\cite{you2021identity} & $\checkmark$ & $\checkmark$ & more powerful than 1-WL \\ \cline{2-5} 
 & Qian et al.~\cite{qian2022ordered} &  & $\checkmark$ & incomparable to k-WL \\ \cline{2-5} 
 & Zhou et al.~\cite{zhou2023relational} & $\checkmark$ & $\checkmark$ & more powerful than k-WL \\ \cline{2-5} 
 & Frasca et al.~\cite{frasca2022understanding} &  & $\checkmark$ & not more powerful than 3-WL \\ \cline{2-5} 
 & Huang et al.~\cite{huang2022boosting} & $\checkmark$ &  & more powerful than k-WL \\ \cline{2-5} 
 & Tahmasebi et al.~\cite{tahmasebi2023power} &  & $\checkmark$ & more powerful than k-WL \\ \hline
\multirow{4}{*}{Non-equivariant GNNs} & Murphy~\cite{murphy2019relational} & $\checkmark$ & $\checkmark$ & more powerful than 1-WL \\ \cline{2-5} 
 & Sato et al.~\cite{sato2021random} & $\checkmark$ & $\checkmark$ & more powerful than 1-WL \\ \cline{2-5} 
 & Abboud et al.~\cite{abboud2021surprising} & $\checkmark$ &  & more powerful than 1-WL \\ \cline{2-5} 
 & Sato et al.~\cite{sato2019approximation} &  & $\checkmark$ & more powerful than 1-WL \\ \hline
\multirow{5}{*}{Others} & Balcilar et al.~\cite{balcilar2021breaking} &  & $\checkmark$ & more powerful than 1-WL \\ \cline{2-5} 
 & Dimitrov et al.~\cite{dimitrov2024plane} & $\checkmark$ & $\checkmark$ & not more powerful than 3-WL \\ \cline{2-5} 
 & Beaini et al.~\cite{beaini2021directional} & $\checkmark$ & $\checkmark$ & more powerful than 1-WL \\ \cline{2-5} 
 & Vignac et al.~\cite{vignac2020building} & $\checkmark$ &  & more powerful than 1-WL \\ \cline{2-5} 
  & Puny et al.~\cite{puny2023equivariant} & $\checkmark$ & $\checkmark$ & more powerful than 3-WL \\ \cline{2-5} 
 & Barcelo et al.~\cite{barcelo2020expressive} &  & $\checkmark$ & more powerful than 1-WL \\ \hline
 \end{tabular}
 \begin{tablenotes}
        \small
        \item[*] {Expressive power w.r.t. WL hierarchy}
    \end{tablenotes}
\end{threeparttable}
}
\end{table*}
\subsection{Discussion}
In this section, we have reviewed the theory of the expressive power of GNNs from multiple perspectives and categorized the methods to devise GNN architectures that are more powerful than 1-WL. Since the hierarchy of the WL algorithm for graph isomorphism problem is the mainstream measurement to characterize the expressive power of different GNN models, we summarize the expressive power of existing GNN architectures in terms of the WL hierarchy and the corresponding techniques in Table~\ref{Table:expressive_power}. After that, we spotlight and discuss four possible future directions on the theory of the expressive power of GNNs.

\textbf{Break the limitation of WL algorithm.} Although the hierarchy of WL algorithm has been prevalent for characterizing the expressive power of GNN in the past few years, the limitations of it has drawn more and more attention recently. On the one hand, the WL algorithm fails to measure the degree of similarity between non-isomorphic graphs due to the binary output for the graph isomorphism problem. Therefore, a more fine-grained metric for expressive power is expected. On the other hand, the WL algorithm is suspected of having the ability to represent the true expressive power of GNN model as it is demonstrated empirically that the more expressive GNN with respect to WL hierarchy do not necessarily have better performance on real world tasks~\cite{velivckovic2022message}. As mentioned before, the WL algorithm neglects some graph properties such as distance between nodes, and thus can leave out potentially important structural information, which make WL algorithm not suitable for other interested graph-related tasks in real world. Hence, a metric with practical value is desirable. 

However, both of these above-mentioned two points are very challenging from the perspective of theory since it requires a transition from qualitative to quantitative and the knowledge of graph theory should be considered to grasp the essential expressive power across multifarious tasks and graphs. There are still some notable attempts. To derive a more fine-grained metric, approximate isomorphism~\cite{kriege2018property} that quantifies the similarity by some graph distance metric is put forward and Boker et al.~\cite{boker2024fine} propose continuous extension of both 1-WL and MPNNs to graphons via evaluating a specific graph metrics on graphons, which is capable of subtly representing the ability to capture the similar graph structure. Notably, Zhang el al.~\cite{zhang2024beyond} derive a novel expressive measure termed homomorphism expressivity based on substructure counting under homomorphism, which provides a quantitative and practical framework to solve both two issues. Furthermore, the proposed novel expressive power hierarchy is closed related to the polynomial expressive power hierarchy proposed by Puny et al.~\cite{puny2023equivariant} that also mitigates the limitation of WL hierarchy.

\textbf{Expressive power for node classification and link prediction.} Due to the dominant WL algorithm for graph isomorphism problem, the majority of introduced works focus on the graph classification task. While node classification and link prediction are two fundamental tasks in graph learning, it is meaningful to analyze the expressive power of GNNs with respect to them in theory. A direct thought inspired by WL algorithm is to distinguish nodes and potential links that can not be determined by the 1-WL algorithm, which can be resolved through techniques discussed previously such as node-based subgraph extraction and node identifier. To characterize the expressive power of GNNs for node classification and link prediction tasks further, one can derive a novel version of WL. 

For example, Barcelo et al.~\cite{barcelo2022weisfeiler} propose the relational WL algorithm to study the expressive power of architectures for link prediction over knowledge graphs. Later, Huang et al.~\cite{huang2024theory} generalized the results to a wider range of models and designed a framework called C-MPNN whose expressive power can be characterized by both the relational WL algorithm and first-order logic. Hu et al.~\cite{hu2022two} discusses the link expressive power of a series of GNNs based 2-WL, 2-FWL and their local variants. Zhang et al.~\cite{zhang2021labeling} reveal the fundamental expressivity limitation of combining two node representations obtained by a GNN as a link representation and propose a labeling trick to enhance GNN's link expressive power. 

For the node classification task, the similarity between different nodes should be taken into consideration and a phenomenon called over-smoothing that the feature of nodes become indistinguishable in deep GNNs has gained much attention. We will introduce it in detail in Section~\ref{sec:exp_long_range}.

\textbf{Expressive power of graph transformer.} The graph transformer is a popular topic in graph representation learning in past few years for producing SOTA results on several graph benchmarks, especially the tasks requiring long-range dependency. To delve into the great success of graph transformers, researchers have studied the expressive power of graph transformers extensively. One attempt is to establish the connection between graph transformers and MPNN. Kim et al.~\cite{kim2022pure} prove that the graph transformer with appropriate positional encoding can approximate any linear permutation-equivariant operators and thus is not less powerful than $k$-IGN and $k$-WL and strictly more powerful than MPNN. Conversely, Cai et al.~\cite{cai2023connection} show that by augmenting the input graph with a virtual node connecting to all graph nodes, MPNN can simulate certain type of graph transformer under mild assumptions. 

Although the connection demonstrates the powerful expressivity of graph transformer, the general transformer architecture does not have an advantage over GNN architecture in terms of expressive power since it is permutation-equivariant and thus fails to distinguish nodes with different positions. More specifically, Zhou et al.~\cite{zhou2024theoretical} prove that the $k$-order graph transformers operating on $k$-order tensors without the positional encoding to structural information are strictly less expressive than $k$-WL. Therefore, more efforts are made to study and utilize the additional expressive power brought by positional encoding and structural encoding. 

For instance, Black et al.~\cite{black2024comparing} compare the expressive power of Absolute Positional Encoding(APE) and Relative Positional Encoding(RPE) in terms of distinguishing non-isomorphic graphs by introducing a variant of WL algorithm. With this framework, the authors prove that the two types of positional encoding are equivalent in distinguishing non-isomorphic graphs and further provide an approach to convert the positional encoding to each other while maintaining the expressive power. Similarly, Zhu et al.~\cite{zhu2023structural} also propose a novel WL algorithm named SEG-WL based on structural encoding to characterize the expressive power of graph transformers. One notable position encoding is the eigenfunction of Laplacian. Equipped with the positional encoding, the graph transformers can go beyond 1-WL~\cite{kreuzer2021rethinking,dwivedi2021graph} and further proved to be universal~\cite{kreuzer2021rethinking}. However, such positional encodings are not permutation-equivariant thus recent works attempt to design equivariant Laplacian positional encoding~\cite{lim2022sign,wang2022equivariant}. 

\textbf{Expressive power of GNNs on specific graphs.} Besides the common graph represented by $(V,E,X)$, many graphs arising in real applications have additional properties and constraints thus corresponding GNN architectures are proposed to handle the tasks on them. To provide a theoretical guarantee for applying the architectures on real-world tasks and further improve the performance, the characterization of expressive power of GNNs on the specific graphs is instrumental since the original WL hierarchy does not take the distinction of specific graphs into account. For geometric graphs that are widely used to represent a 3D atomic system, Joshi et al.~\cite{joshi2023expressive} propose a geometric version of WL(GWL) by considering geometric graph isomorphism that requires underlying graphs to be not only topologically isomorphic but also equivalent concerning some symmetry groups of permutation, translation, rotation, and reflection. Using the framework, the authors analyze the impact of some key factors of geometric GNNs including depth, tensor order, and body order on the expressive power and further derive the equivalence between the geometric graph isomorphism test and universal approximation ability of geometric GNNs~\cite{dym2020universality,villar2021scalars}. 

Different from Joshi's work, Beddar et al.~\cite{beddar2024weisfeiler} extend the 1-WL to attribute and dynamic graphs and further establish the connection between the novel version of WL algorithm and unfolding trees. Besides, the expressive power of GNNs on relational graphs has been discussed before. In addition, it is observed that existing works often try to generalize the WL framework to specific graphs, which may not be suitable for the real application as mentioned before. Therefore, generalization of other existing measurements such as subgraph counting or characterization via distinct tasks on corresponding graphs is worth exploring.

\section{Generalization}
\label{sec:gen}
Generalization refers to the ability of a hypothesis or a learning algorithm to work well on unseen data, which is one of the most critical perspectives of machine learning algorithms. To quantitatively analyze the generalization property, the generalization (error) bound provides a theoretical guarantee and has drawn much attention in deep learning. 

In this paper, we focus on the generalization bound on graph learning and provide a systematic analysis. Specifically, we consider the generalization bound of GNNs for graph classification task and node classification task. Although the dependent and unstructured property of graph data and the complex design of graph neural networks pose difficulty for deriving the generalization bound of GNNs precisely, the framework and methods developed in deep learning theory can facilitate the computation and provide insightful generalization bound with some assumptions and simplification. Consistent with the conventions in deep learning, we classify the literature into four groups: the complexity of hypothesis space-based, PAC-Bayes based, stability-based, and graph neural tangent kernel (GNTK) based, with GNTK serving as an extension of the neural tangent kernel (NTK). Notably, the primary disparity lies in how the generalization bound on graphs integrates the statistical characteristics of graphs and the learning matrix of Graph Neural Networks (GNNs). Subsequent subsections will detail these categories individually.

\subsection{Notations and problem formulation}
Before delving into the concrete methods to derive generalization bounds of GNN, we first introduce some necessary notations and provide the formulation of the problem. Consider the training dataset $Z=\{(x_1,y_1),(x_2,y_x)\ldots,(x_m,y_m)\}$ with $m$ samples where $x_i\in \mathcal{X}$ is the feature and $y_i\in \mathcal{Y}$ is the label. All $(x_i,y_i)$ are i.i.d. observed from an underlying distribution $\mathcal{D}$ over $\mathcal{X}\times\mathcal{Y}$. Then the learning algorithm $A$ attempts to learn hypothesis $h:\mathcal{X}\rightarrow\mathcal{Y}$ from the training dataset and the hypothesis space $\mathcal{H}$ consists of all possible hypothesis $h$. For any hypothesis $h$ learned by an algorithm $A$ from the training dataset $Z$, the empirical risk $\widehat{\mathcal{R}}(h)$ and expected risk $\mathcal{R}(h)$ with respect a loss function $\ell$ are defined respectively as follows,
\begin{definition}[Empirical risk and expected risk]
    \begin{equation}
    \widehat{\mathcal{R}}_Z(h)=\frac{1}{m}\sum_{i=1}^m\ell(h,(x_i,y_i)),\mathcal{R}(h)=\mathbb{E}_{(x,y)\sim \mathcal{D}}\ell(h,(x,y)).
\end{equation}
\end{definition}

In addition, when algorithm $A$ is a randomized algorithm, we consider the hypothesis $A_Z$ learned by $A$ from the training dataset $Z$ and compute the expectation of the empirical risk $\widehat{\mathcal{R}}(A_Z)$ and expected risk $\mathcal{R}(A_Z)$ with respect to the randomness introduced by $A$ respectively as follows,
\begin{definition}[Empirical risk and expected risk in randomized algorithm]
    \begin{equation}
\widehat{\mathcal{R}}_S(A)=\mathbb{E}_A[\widehat{\mathcal{R}}_S(A_Z)],\mathcal{R}(A)=\mathbb{E}_A[\mathcal{R}(A_Z)].
\end{equation}
\end{definition}

Then, the generalization bound can be analyzed by the generalization gap $\mathcal{R}(h)-\widehat{\mathcal{R}}(h)$. Besides, it is noted that we only consider the in-distribution generalization bound since it is the common assumption for most of the introduced methods and we will further cover the out-of-distribution generalization in the discussion.

\subsection{Complexity of hypothesis space based}

Given that the learning algorithm is situated within a hypothesis space, the complexity of this space plays a crucial role in defining the range of problems the algorithm can address. Therefore, it is common practice to analyze the complexity of the hypothesis space in order to establish a generalization bound, which can be assessed using theoretical measures such as Vapnik-Chervonenkis dimension (VC-dim) ~\cite{vapnik2015uniform} and Rademacher complexity~\cite{bartlett2002rademacher,koltchinskii2001rademacher}, along with covering numbers~\cite{dudley1967sizes}. Here we only introduce the first two methods that have been used to obtain the generalization bound of GNN as the covering number often acts as an alternative of Rademacher complexity and is applied to more complex settings in the theory of deep learning.

Note that the VC-dim~\cite{vapnik2015uniform} for binary classification task is defined upon growth function. Therefore, we will first introduce the definition of the growth function before discussing the generalization guarantee associated with VC dimension.

\begin{definition}[Growth function]
   For any non-negative integer m, the growth function of hypothesis space $\mathcal{H}$ is defined as follows:
\begin{equation}
    \Pi_{\mathcal{H}}(m):=\max\limits_{x_1,\ldots,x_m\in \mathcal{X}}|\{h(x_1),\ldots,h(x_m):h\in\mathcal{H}\}|.
\end{equation} 
\end{definition}

The growth function represents the maximum number of possible labeling of $m$ data points by $\mathcal{H}$. If $\Pi_{\mathcal{H}}(m)=2^m$ which means any labeling of $m$ data points can be perfectly classified by one hypothesis in the hypothesis space, we say $\mathcal{H}$ shatters the dataset $\{x_1,\ldots,x_m\in \mathcal{X}\}$, and the VC-dim of $\mathcal{H}$ is defined as the largest $m$. From the above definition, the VC dim is independent of the data distribution thus a more universal approach. The generalization bound based on VC-dim can be obtained by the following theorem.
\begin{theorem}[Generalization guarantee of VC-dim] Assume hypothesis space $\mathcal{H}$ has VC-dim $D$, $m$ is the training set size. Then, for any $\delta>0$, with probability $1-\delta$, the following inequality holds for any $h\in\mathcal{H}$,
\begin{equation}
    \mathcal{R}(h)\leq \widehat{\mathcal{R}}(h)+\sqrt{\frac{2D\log\frac{em}{D}}{m}}+\sqrt{\frac{\log\frac{1}{\delta}}{2m}}.
\end{equation}
\end{theorem}
 
Bounding the VC-dim by $O(p^4N^2)$ where $p$ is the number of parameters in GNN and $N$ is the size of input nodes, Scarselli~\cite{scarselli2018vapnik} derives generalization bounds of node classification task on graphs for the first time. The result suggests that generalization ability improves with the increasing number of nodes and parameters. Besides, it is worth noting that the bound is identical to that of recurrent neural networks.

In contrast to VC-dimension, Rademacher complexity ~\cite{bartlett2002rademacher,koltchinskii2001rademacher} takes into account the distribution of data and provides a more nuanced assessment of the richness of the hypothesis space by quantifying how well the hypothesis set can accommodate random noise. The empirical definition of Rademacher complexity is outlined below.
\begin{definition}[Empirical Rademacher complexity]
    Given a function class $\mathcal{H}$ and a dataset $Z$ with $m$ samples $Z=\{x_1,\ldots,x_m\}$, the empirical Rademacher complexity of $\mathcal{H}$ can be defined by:
\begin{equation}  
    \hat{\mathfrak{R}}_m(\mathcal{H}) = \mathbb{E}_{\chi}[\sup\limits_{h\in\mathcal{H}}\frac{1}{m}\sum_{i=1}^{m}\chi_ih(x_i)],
\end{equation}
where $\chi=\{\chi_1,\ldots,\chi_m\}$ is a random vector whose items are uniformly chosen form $\{-1,+1\}$ and samples are i.i.d. generated from a distribution $\mathcal{D}$.
\end{definition}
Further, the Rademacher complexity of $\mathcal{H}$ is defined as $\mathfrak{R}_m(\mathcal{H}) = \mathbb{E}_{S\sim\mathcal{D}^m}{\widehat{\mathfrak{R}}_m(\mathcal{H})}$. The (empirical) Rademacher complexity can deduce the generalization bound on binary classification and regression tasks according to the following theorem.
\begin{theorem}[Generalization guarantee of Rademacher complexity] Given a function class $\mathcal{H}$ containing functions $h:\mathcal{X}\rightarrow[a,b]$ and a dataset $S$ with $m$ samples $S=\{x_1,\ldots,x_m\}$. Then for any $\delta>0$ and $h \in \mathcal{H}$, with probability $1-\delta$, we have,
\begin{align}
    \mathbb{E}[h(x)]\leq\frac{1}{m}\sum_{i=1}^{m}h(x_i)+2\mathfrak{R}_m(\mathcal{H})+(b-a)\sqrt{\frac{\ln\frac{1}{\delta}}{2m}}, \\
    \mathbb{E}[h(x)]\leq\frac{1}{m}\sum_{i=1}^{m}h(x_i)+2\widehat{\mathfrak{R}}_m(\mathcal{H})+3(b-a)\sqrt{\frac{\ln\frac{2}{\delta}}{2m}}.
\end{align}
\end{theorem}

In order to apply the theorem to the graph classification task that calculates the average of binary prediction on each node to obtain the graph label, Garg et al.~\cite{garg2020generalization} bound the Rademacher complexity of GNN by considering the Rademacher complexity of the computation tree of each node whose update formula of GNN follows a mean field form:
\begin{equation}
    x_v^l=\phi(\mathbf{W}_1x_v+\mathbf{W}_2\rho (\sum_{u\in \mathcal{N}(v)}g(x_u^{l-1}))). 
\end{equation}

To illustrate the varying generalization bound with respect to the different parameters, they define a combination parameter $\mathcal{C}=C_{\rho}C_gC_{\phi}B_{\mathbf{W}}$ that is the product of the Lipschitz constant of $\rho,g,\phi$ and $B_{\mathbf{W}}$ denotes the bound norm of weight $\mathbf{W}_2$. Let $d$ denote the maximum degree of nodes in the graph, $r$ denote the dimension of embedding, $L$ denote the number of layers, $m$ denote the size of training nodes and $\gamma$ denote margin parameter in the loss, the dependency can be expressed as follows:
\begin{equation}
    \begin{cases}
    O(\frac{rd}{\sqrt{m}\gamma}) & \text{for} \quad \mathcal{C}<\frac{1}{d} \\
    O(\frac{rdL}{\sqrt{m}\gamma})  & \text{for} \quad \mathcal{C}=\frac{1}{d} \\
    O(\frac{rd\sqrt{rL}}{\sqrt{m}\gamma}) & \text{for} \quad \mathcal{C}>\frac{1}{d}
    \end{cases}
    \label{eq:dependency}
\end{equation}

The generalization bound given by Equation (\ref{eq:dependency}) is much tighter than the counterpart of VC-dim since the latter has a higher order dependency on the size of parameters of the neural network and the size of input $N$ is at least the maximum degree $d$. In addition, they find that the bound for GNN in the equation is comparable to that of RNN, which indicates that GNN can be seen as the sequentialized RNN. It is noticed that the observation is consistent with that of the generalization bound obtained by VC-dim. Besides, Lv~\cite{lv2021generalization} also proves a generalization bound by Rademacher complexity of GCN with one hidden layer on the node classification task. The derived bound is sharp for having a lower bound matching the upper bound.

In contrast to above-mentioned works which focus on generalization bounds for GNNs in the inductive setting, there is a growing body of research examining generalization bounds in the more realistic transductive setting. These studies typically leverage transductive Rademacher complexity~\cite{el2009transductive} to derive generalization bound, which considers unobserved samples when deriving generalization bounds. Formally, denote the size of training set and test set as $m$ and $u$ respectively, the target is to learn a function that generates the best predictions for the label of the test set based on the features for both training and test set and the labels for the training set. Within the transductive framework, Oono and Suzuki~\cite{oono2020optimization}  establish a generalization bound for multi-scale GNNs in node classification tasks, while Esser et al.~\cite{esser2021learning} refine the bound for multi-layer GNNs within a planted model to investigate graph and feature alignment. In addition, Deng et al.~\cite{deng2022graph} provide a generalization upper bound to guarantee to performance of GNN based recommendation systems. Recently, Tang and Liu~\cite{tang2023towards} introduce a high probability generalization bound for GNNs trained with SGD in the transductive setting, accounting for the comparable performance of shallow and deep models, as well as the effectiveness of techniques like early stopping and dropedge to some extent.

\subsection{PAC-Bayes based}
Over recent years, the probably approximately correct (PAC)-Bayesian approach~\cite{mcallester1999pac,langford2002pac} has garnered significant attention for providing a more realistic and tighter generalization bound compared to traditional VC-dimension-based and Rademacher complexity-based bounds. Originally introduced to measure the learnability of a problem, the PAC concept~\cite{valiant1984theory} assesses whether a learning algorithm can output the correct result with high probability when given a specific number of training examples from an unknown distribution. Since the bounds provided by classical PAC framework is unsatisfactory due to the large hypothesis space, the PAC-Bayesian approach incorporates the Bayesian view by putting a prior distribution over the hypothesis space to derive tighter generalization bounds for the target machine learning model. In the following discussion, we will present the framework developed by PAC-Bayesian theory for analyzing the generalization bound of GNNs.

In this context, we focus on the margin bound, utilizing a multi-class margin loss function with a threshold $\gamma$ utilized. The original empirical loss is defined as:

\begin{equation}
    \ell_{Z,\gamma}(h)=\frac{1}{m}\sum_{i=1}^m\mathbf{1}\left[h(x_i)[y_i]\leq \max_{j\neq y_i}h(x_i)[j]+\gamma\right],
\end{equation}
where $Z$ is the training set with $m$ examples. Further, the original generalization loss is given as:
\begin{equation}
    \ell_{\mathcal{D},\gamma}(h)=Pr_{x\sim \mathcal{D}}\left[h(x)[y]\leq \max_{j\neq y}h(x)[j]+\gamma\right],
\end{equation}
where $\mathcal{D}$ is an unknown distribution from which samples are generated.

Assume there is a prior distribution $\mathcal{P}$ and a posterior distribution $\mathcal{Q}$ over the model parameter $\theta$ in Bayesian theory, then the empirical loss and generalization loss are defined by expectation which are denoted as $\mathbb{E}_{\theta\sim \mathcal{Q}}[\widehat{\mathcal{R}}_{\gamma}(h(\theta))]$ and $\mathbb{E}_{\theta\sim \mathcal{Q}}[\mathcal{R}_{\gamma}(h(\theta))]$ respectively. The generalization bound of the model based on PAC-Bayesian can be obtained via the following theorem.
\begin{theorem}[Generalization guarantee of PAC-Bayesian]
~\cite{mcallester2003simplified}: Let $\mathcal{P}$ and $\mathcal{Q}$ be the prior distribution and posterior distribution over the model with parameter $\theta$, and $Z$ be a dataset with $m$ samples generated i.i.d. from distribution $\mathcal{D}$. Then, for any $\delta \in (0,1)$ and $h \in \mathcal{H}$, with probability $1-\delta$, we have
\begin{equation}
    \mathbb{E}_{\theta\sim \mathcal{Q}}[\mathcal{R}_{\gamma}(h(\theta))] \leq \mathbb{E}_{\theta\sim\mathcal{Q}}[\widehat{\mathcal{R}}_{\gamma}(h(\theta))]+\sqrt{\frac{\mathbf{KL}(\mathcal{Q}||\mathcal{P})+\log\frac{m}{\delta}}{2(m-1)}}.
    \label{the:pac}
\end{equation}
\end{theorem}

In Theorem~\ref{the:pac}, the selection of distribution pairs 
 $\mathcal{P}$ and $\mathcal{Q}$ can be arbitrary. However, choosing disparate distributions may render the computation of KL-divergence challenging, while oversimplified choices could result in significant empirical and generalization losses. To address this issue, Neyshabur et al.~\cite{neyshabur2017pac} offer an effective approach to determine the bound by considering the posterior distribution over parameters as a perturbation distribution derived from two known distributions. The perturbation-based Bayesian generalization bound can be formally defined as follows:
\begin{theorem}[Perturbation based Bayesian generalization bound] Let $h\in \mathcal{H}:\mathcal{X}\rightarrow \mathbb{R}^K$ be any model with parameter $\theta$, $S$ be the dataset with $m$ samples that generated i.i.d. from distribution $\mathcal{D}$, and $\mathcal{P}$ be the prior distribution on the parameters that is independent of the training data. Then, for any $\gamma,\delta>0$, any parameter $\theta$ and any random perturbation $\Delta\theta$ s.t. $Pr_{\phi}\left[\max_{x\in \mathcal{X}}|h(\theta+\Delta\theta)-h(\theta)|_{\infty}< \frac{\gamma}{4}\right]> \frac{1}{2}$, with probability at least $1-\delta$, we have:
\begin{equation}
    \ell_{\mathcal{D},0}(h(\theta)) \leq \ell_{Z,\gamma}(h(\theta))+\sqrt{\frac{2\mathbf{KL}(Q(\theta+\Delta\theta)||\mathcal{P})+\log\frac{8m}{\delta}}{2(m-1)}}.
\end{equation} 
\label{the:Bayesian_gb}
\end{theorem}

As suggested by Theorem~\ref{the:Bayesian_gb}, if a model's output remains stable for any input even after a slight parameter perturbation with high probability, the generalization bound can be established by bounding the output change under perturbation. Inspired by this framework, Liao et al.~\cite{liao2020pac} first apply PAC-Bayesian approach to GNNs and utilize the above theorem to derive the generalization bounds for GCNs and MPGNNs on graph classification tasks. The perturbation analysis involved controlling the maximum node representation and the maximum change of node representation. Here, we present the generalization bound for GCNs, with the prior distribution  $\mathcal{P}$  and the perturbation distribution modeled as a Gaussian distribution with zero mean and covariance matrix:

\begin{theorem}[PAC-Bayesian Generalization bound for GCNs] For any $B>0$, $L>1$, $d>1$, $r>1$, let $f\in \mathcal{H}:\mathcal{X}\times \mathcal{G}\rightarrow \mathbb{R}^K$ be a $L$-layer GCN and $Z$ be a dataset with $m$ samples that generated i.i.d. from distribution $\mathcal{D}$. Then for any $\delta,\gamma>0$, with probability at least $1-\delta$, we have,
\begin{equation}
\begin{aligned}
    \ell_{\mathcal{D},0}(h(\theta)) &\leq \ell_{Z,\gamma}(h(\theta))\\ & +\mathcal{O}\left(\sqrt{\frac{B^2d^{L-1}L^2r\log(Lr) \mathbf{W}+\log\frac{ml}{\delta}}{\gamma^2m}}\right),
\end{aligned}
\end{equation}
where
$\mathbf{W} =\prod_{i=1}^l||\mathbf{W}_i||_2^2\sum_{i=1}^l\frac{||\mathbf{W}_i||_F^2}{||\mathbf{W}_i||_2^2}$, $B$ is upper bound for the $l_2$ norm of the feature, $L$ is the number of layers, $d$ is the maximum node degree considering itself, $r$ is the maximum hidden dimension, $W_i$ is the weight matrix for the $i$-th layer.
\end{theorem}
The bound implies a dependency on the maximum node degree $d$, maximum hidden dimension $r$, and spectral norm of weight matrix. Compared to the previous Rademacher complexity bound for MPNN, the PAC-Bayesian generalization bound is more tight with respect to the maximum node degree $d$ and maximum hidden dimension $r$. Specifically, for maximum node degree the PAC-Bayesian bound scale as $\mathcal{O}(d^{L-1})$ while Rademacher complexity bound scale as $\mathcal{O}\left(d^{L-1}\sqrt{\log(d^{2L-3})}\right)$. For maximum hidden dimension the PAC-Bayesian bound scale as $\mathcal{O}\left(\sqrt{r\log r}\right)$ while Rademacher complexity bound scales as $\mathcal{O}\left(r\sqrt{\log r}\right)$. 

The comparison of dependency on spectral norm of weight is inaccessible theoretically without knowing the actual value. It is also noticeable that the maximum node degree is the only graph statistics factor in the generalization bound, which indicates that the relationship between graph structure and generalization ability may not be fully explored. Following Liao's work, Sales~\cite{sales2022generalization} further improve the bound by reducing the factor of the exponential term of maximum node degree and utilizing a theorem on random matrix to bound the spectral norm more precisely.

Ju et al.~\cite{ju2023generalization} delves into the relationship between the generalization bound and the graph diffusion matrix, which offers a more detailed representation of the graph structure. By refining the perturbation analysis using Hessians, they achieve a tighter and more precise bound that scales with the largest singular value of the diffusion matrix instead of the previous method based on the maximum node degree. In a separate study, Sun and Lin~\cite{sun2024pac} apply the PAC-Bayesian framework to the adversarial robustness setting. They derive adversarially robust generalization bounds for both GCNs and MPNNs in graph classification tasks. This new bound eliminates the exponential dependency on the maximum node degree.

\subsection{Stability based}
Besides the perturbation analysis that exerts perturbation to the weight matrix in PAC-Bayesian approach to guarantee the generalization, it is intuitive that the performance of an algorithm with good generalization ability does not degrade much after a slight change in the training data. The stability is the measurement to quantify the change of the output of an algorithm when the training data is modified. Here we only introduce the uniform stability~\cite{bousquet2002stability} that is most widely-used in deriving the generalization bound. Before giving the formal definition of the uniform stability, we introduce modification operation to the training data in advance.

\textbf{Data Modification}~\cite{bousquet2002stability} Let $\mathcal{X}$ be the input space, $\mathcal{Y}$ be the output space and $\mathcal{Z}=\mathcal{X} \times \mathcal{Y}$. For $x_i\in\mathcal{X}$ and $y_i\in\mathcal{Y}\subset\mathbb{R}$, let $Z$ be a training set with m examples $Z=\{z_1=(x_1,y_1),\ldots,z_m=(x_m,y_m)\}$ and all samples are i.i.d. from $\mathcal{D}$. Two fundamental modifications to the training set $Z$ are as follows: 
\begin{itemize}
    \item Removing $i^{th}$ data point in the set $Z$ is represented as,
\begin{equation}
    Z^{\backslash i}={z_1,\ldots,z_{i-1},z_{i+1},\ldots,z_m},
\end{equation}

    \item Replacing $i^{th}$ data point in the set $Z$ is represented as,
\begin{equation}
    Z^i={z_1,\ldots,z_{i-1},z_i^{'},z_{i+1},\ldots,z_m}.
\end{equation}
\end{itemize}
Then the uniform stability for a randomized algorithm can be defined as follows,
\begin{definition}[Uniform stability]
Let $A$ be a randomized algorithm trained on dataset $S$ and $A_Z$ is the output hypothesis, then $A$ is $\beta_m$-uniformly stable with respect to a loss function $\ell$, if it satisfies,
\begin{equation}    \sup\limits_{Z,z}|\mathbb{E}_A[\ell(A_Z,z)]-\mathbb{E}_A[\ell(A_{Z^{\backslash i}},z)]|\leq \beta_m.
\end{equation}
\end{definition}

The request for uniform stability is strict since it holds for every possible training set with $m$ samples. In addition, a randomized algorithm is taken into consideration to analyze the model optimized by some randomized algorithm e.g. stochastic gradient descent (SGD). Then, the generalization gap based on the uniform stability can be derived by the following Theorem.
\begin{theorem}[Generalization guarantee of uniform stability] Assume a uniformly stable randomized algorithm $(A_Z,\beta_m)$ with a bound loss function $0\leq \ell(A_Z,z)\leq M$ for any $Z,z$. Then, for any $\delta>0$, with probability $1-\delta$ over choice of an i.i.d size-m training set $Z$, we have:
\begin{equation}
    \mathbb{E}[\mathcal{R}(A_Z)-\widehat{\mathcal{R}}(A_Z)]\leq2\beta_m+(4m\beta_m+M)\sqrt{\frac{\log\frac{1}{\delta}}{2m}}.
\end{equation}
\label{th:uniform stab}
\end{theorem}

With Theorem \ref{th:uniform stab}, the generalization bound can be obtained via proving the uniform stability of an algorithm. Besides, to ensure that the generalization gap can converge to 0, it requires $\beta_m$ to decay faster than $\mathcal{O}(\frac{1}{\sqrt{m}})$ as $m\rightarrow\infty$. 

Assuming the randomized algorithm to be a single-layer GCN optimized by SGD, Verma and Zhang~\cite{verma2019stability} first analyze the uniform stability of GCN and derives the stability-based bound on the node classification task. They derive the uniform stability constant and corresponding generalization bound of the model, which are given in Theorem \ref{the:uniform_stab} and Theorem \ref{the:gb_gcn}, respectively.
\begin{theorem}[Uniform stability of GCN using SGD]\label{the:uniform_stab} Let the loss and activation function be Lipschitz-continuous and smooth functions. Then a single layer GCN model training by SGD algorithm for $T$ iterations is $\beta_m$-uniformly stable, where
    \begin{equation}
        \beta_m \leq\left(\eta \alpha_{\ell} \alpha_\sigma v_{\ell}\left(\lambda_{\mathcal{G}}^{\max }\right)^2 \sum_{t=1}^T\left(1+\eta v_{\ell} v_\sigma\left(\lambda_{\mathcal{G}}^{\max }\right)^2\right)^{t-1}\right) / m, 
    \end{equation}
where $\eta>0$ is the learning rate, $\alpha_{\ell},\alpha_\sigma>0$ are Lipschitz constants for the loss function and activation function respectively, $v_{\ell},v_\sigma>0$ are the Lipschitz constants for the gradient of loss function and activation function respectively, and $\lambda_G^{\max }$ is the largest absolute eigenvalue of the graph diffusion matrix.
\end{theorem}
\begin{theorem}[Generalization bound for GCN training by SGD]\label{the:gb_gcn} Let the loss and activation function be Lipschitz-continuous and smooth functions, then a single layer GCN model training by SGD algorithm for $T$ iterations is $\beta_m$-uniformly stable. The generalization bound based on uniform stability is given as follows,
    \begin{equation}
    \begin{aligned}
        \mathbb{E}_{SGD}[\mathcal{R}(A_Z)-\hat{\mathcal{R}}(A_Z)]&\leq  \frac{1}{m}\mathcal{O}((\lambda_{\mathcal{G}}^{\max})^{2T}) \\ & + (\mathcal{O}((\lambda_{\mathcal{G}}^{\max})^{2T})+M)\sqrt{\frac{\log\frac{1}{\delta}}{2m}}
    \end{aligned}
    \end{equation}
\end{theorem}

The obtained uniform stability constant and generalization bound make sense since the largest absolute eigenvalue of graph diffusion matrix can be controlled by various normalization methods, which is in accord with the stable training behavior and better performance when adopting a normalized graph diffusion matrix. They also stress the importance of batch-normalization that has similar effect on the training of multi-layer GNNs. However, the bound is incomparable to those of previous for considering a randomized learning algorithm. Later, Zhou and Wang~\cite{zhou2021generalization} extend the work to multi-layer GNNs and further demonstrate that increasing number of layers can enlarge the generalization gap. 

Different from Zhang~\cite{verma2019stability}'s setting, Cong et al.~\cite{cong2021provable} consider a transductive setting and study the generalization bound of multi-layer GCN optimized by full batch gradient descent on node classification task by transductive uniform stability. They delve into the Lipschitz continuity, smoothness, and gradient scale to compare the generalization bound of different models.

\subsection{GNTK based}
Neural tangent kernel (NTK)~\cite{arora2019fine,jacot2018neural} is a kernel-based method to analyze over-parameterized neural networks trained by gradient descent in the infinite-width limit in deep learning. Du et al.~\cite{du2019graph} generalize the theory to graph learning via combining GNN with graph kernels. Specifically, they consider the graph classification task training on $n$ graphs $\mathcal{G}=\{\mathcal{G}_1,\mathcal{G}_2,\ldots,\mathcal{G}_n\}$. Let $f(\theta, \mathcal{G}_i)$ be the output of the GNN parameterized by $\theta$ testing on graph $\mathcal{G}_i$ and $F(t)=\left(f(\theta, \mathcal{G}_i)\right)_{i=1}^n$, then the training dynamics by gradient descent with infinitesimally small learning rate, i.e. $\frac{d\theta}{dt}=-\nabla \ell(\theta(t))$ considering square loss function $\ell(\theta)=\frac{1}{2}\sum_{i=1}^n(f(\theta,\mathcal{G}_i)-y_i)^2$ follows the formula
\begin{equation}
    \frac{dF}{dt}=-\mathbf{H}(t)(F(t)-y),
\end{equation}
where
\begin{equation}
    \mathbf{H}(t)_{ij}=<\frac{\partial f(\theta(t),\mathcal{G}_i)}{\partial \theta}, \frac{\partial f(\theta(t),\mathcal{G}_j)}{\partial \theta}>.
\end{equation}

Since it is proved that for an over-parameterized neural network, the matrix $\mathbf{H}(t)$ is almost constant regardless of different $t$, the training process can be viewed as a kernel regression problem. To be further, the matrix $\mathbf{H}(0)$ can converge to a deterministic kernel matrix called NTK if the parameters are randomly initialized by Gaussian distribution. This property facilitates the analysis of the generalization bound of GNN as long as the GNN is converted to its GNTK. Denoted the kernel matrix as $\widehat{\mathbf{H}}$. They provide the technique to perform the conversion and derive the generalization bound of a single-layer GNN via Rademacher complexity:
\begin{theorem}[GNTK Generalization bound for GNN] Given $n$ training data $\{(\mathcal{G}_i,y_i)\}_{i=1}^n$ drawn i.i.d. from the underlying distribution $\mathcal{D}$. Then for any loss function $\ell:\mathbb{R}\times \mathbb{R}\rightarrow [0,1]$ that is 1-Lipschitz in the first argument such that $\ell(y,y)=0$ and any $\delta>0$, with probability at least $1-\delta$, the generalization loss of the GNTK predictor can be upper bounded by
\begin{equation}
\begin{aligned}
    \mathbb{E}_{(G,y)\sim \mathcal{D}}&[\ell(f_{ker}(\mathcal{G}),y)] \\ & \leq \mathcal{O}\left(\frac{\sqrt{y^T\widehat{\mathbf{H}}^{-1} y\cdot tr(\widehat{\mathbf{H}})}}{n}+\sqrt{\frac{\log(1/ \delta)}{n}}\right).        
\end{aligned}
\end{equation}
    
\end{theorem}

It is observed that the generalization bound derived by GNTK depends on the label $y$ and kernel matrix $\hat{\mathbf{H}}$ which is data-dependent and different from bounds obtained by other methods. The data-dependent generalization bound is directly related to training samples and thus reflects the property of the data generation process. To provide a more concrete bound, they further bound $y^T\widehat{\mathbf{H}}^{-1} y$ and $tr(\widehat{\mathbf{H}})$ respectively to demonstrate that the GNN can learn the corresponding class of graph labeling function with polynomial number of samples. This is the first sample complexity analysis with respect to the generalization bound of GNN.

\begin{table*}[h]
  \caption{Generalization bounds for GNNs.}
  \label{table:generalization_bounds}
  \centering
  \footnotesize
  \begin{threeparttable}
  \begin{tabular}{ccccccccc}
    \toprule
	Task & Work		& Method			& Generalization bound \\
    \midrule
	 Node classification & Scarselli et al.~\cite{scarselli2018vapnik} 	& VC-dimension	& $\mathcal O \left( \frac{r^3N}{\sqrt{m}}\right)$\\
    \midrule
	Node classification & Garg et al.~\cite{garg2020generalization} 	& Rademacher complexity	& 
     $\widehat{C}$  $ \begin{cases}
   $$ O(\frac{rd}{\sqrt{m}\gamma})$$ & \text{for} $$\quad \mathcal{C}<\frac{1}{d}$$ \\
    $$O(\frac{rdL}{\sqrt{m}\gamma})$$  & \text{for} $$\quad \mathcal{C}=\frac{1}{d}$$ \\
    $$O(\frac{rd\sqrt{rL}}{\sqrt{m}\gamma})$$ & \text{for} $$\quad \mathcal{C}>\frac{1}{d}$$
    \end{cases} $ \tnote{1}
\\
    \midrule
	Node classification & Lv~\cite{lv2021generalization} 	& Rademacher complexity	& $\mathcal O \left(|\lambda^{max}_G|\sqrt{\frac{d}{m}}\sum_{i=1}^{d}\max_{j\in \mathcal{N}(v)}\left|\widehat{A}_{ij}\right|\right)$\\
    \midrule
	Node classifaction & Esser et al.~\cite{esser2021learning} 	& Transductive Rademacher complexity	& $\mathcal O\left( \frac{(m+u)^2}{mu} + \log(m+u)\right)$\\
    \midrule
        Node classification & Oono and Suzuki~\cite{oono2020optimization} 	& Transductive Rademacher complexity	& $\mathcal O\left( \frac{(m+u)^{\frac{3}{2}}}{mu}\right)$\\
    \midrule
	Node classification & Tang and Liu~\cite{tang2023towards} 	& Transductive Rademacher complexity	& $\mathcal O\left( \frac{(m+u)^{\frac{3}{2}}}{mu} \right)$\\
    \midrule
	Graph classification & Liao et al.~\cite{liao2020pac} 	& PAC-Bayes 	& $\mathcal O \left(\frac{d^{L-1}\sqrt{r\log r}}{\sqrt{m}\gamma}\right)$\\
    \midrule
	Node classification & Verma and Zhang~\cite{verma2019stability} & Uniform stability	& $\frac{1}{m}\mathcal{O}((\lambda_{\mathcal{G}}^{\max})^{2T})+ (\mathcal{O}((\lambda_{\mathcal{G}}^{\max})^{2T})+M)\sqrt{\frac{\log\frac{1}{\delta}}{2m}}$\\
    \midrule
	Node classification & Cong et al.~\cite{cong2021provable}	&  Transductive uniform stability	& $\frac1\gamma\mathcal{O}\left(\beta\sqrt{\frac{mu}{m+u}}\right)$ \tnote{2}\\
    \midrule
	Graph classification & Du et al.~\cite{du2019graph} 	& Neural tangent kernel	& $\mathcal{O}\left(\frac{\sqrt{y^T\widehat{\mathbf{H}}^{-1} y\cdot tr(\widehat{\mathbf{H}})}}{n}+\sqrt{\frac{\log(1/ \delta)}{n}}\right)$\\
    \bottomrule
  \end{tabular}
\begin{tablenotes}
    \small
    \item[1] {where $\widehat{C}$ may also contain terms that have dependency on $d$} thus the worst dependency of the generalization bound on $d$ is $\mathcal{O}(d^{L-1}\log d)$
    \item[2] {where $\beta$ is the uniform stability constant w.r.t. different GNN architectures}
\end{tablenotes}
\end{threeparttable}
\end{table*}
\subsection{Discussion}

This section presents four main methods for analyzing the generalization bound of GNNs. These methods typically integrate analytical tools from deep learning theory with GNN architecture and graph structure to derive the generalization bounds. However, due to the limitations in the methods themselves and the simplified setting that is far from real applications such as GNNs with a single hidden layer, most of the bounds have an enormous gap compared to empirical results and provide little insight into the architecture design and training techniques. Furthermore, the derived bounds can contradict to empirical result to some degree. For example, the complexity of the hypothesis space bound increases as the number of parameters becomes larger and the stability-based bound grows with respect to the iterations of optimization. To handle the problems and further improve the generalization bound of GNNs, researchers can leverage recent advances in deep learning theory such as local Rademacher complexity~\cite{bartlett2005local}, marginal-likelihood PAC-Bayes~\cite{valle2020generalization} and $\mathcal{H}$-consistency~\cite{mao2024h}. Besides, it is observed that existing generalization bounds often heavily rely on the number of nodes and maximum node degree as the graph-related term in their final expressions, which is too coarse-grained to capture the complex graph structure information. Therefore, considering additional graph statistics in constructing these bounds could establish a closer link between the generalization bound and the underlying graph structure information, potentially enhancing the understanding of how graph characteristics influence generalization performance. 

It is also noted that existing research in this area can also be categorized based on the task addressed, with a focus on node classification and graph classification tasks. Table~\ref{table:generalization_bounds} summarize the generalization error bound of GNNs with respect to the task and the method to derive the bound. Concretely, we only preserve the terms that are related to the input graphs in order to show its dependence on graph properties concisely. Link prediction tasks have been less explored due to challenges related to edge partitioning during training and the complexity of the optional prediction function involving two nodes, which violates the settings that samples are independent.  

Recent studies on the generalization bound of GNNs have shifted towards transductive learning. The transductive learning acknowledges the presence of unlabeled data, which mirrors real-world scenarios in graph-based learning and allows for the integration of optimization algorithms and training data size into the generalization bound. Intuitively, the transductive learning setting shrinks the hypothesis space by attempting to limit the hypothesis space to the space around optimal hypothesis, which provides a tighter and more practical bound compared to the common VC-dim based and Rademacher complexity generalization bounds that consider the complexity of the whole hypothesis space. Therefore, the transductive generalization gap, defined as the difference between training error and testing error, offers a clearer verification of results through experimentation. In addition, methods to derive generalization bounds established in the inductive setting have been extended to the transductive counterparts non-trivially in deep learning theory, such as transductive PAC-Bayesian~\cite{begin2014pac} and transductive Rademacher complexity~\cite{el2009transductive}, hold promise for analyzing the generalization bound of GNNs in transductive learning settings in the future.
 
To deepen our understanding of GNNs and graphs, future research efforts could focus on establishing generalization bounds for GNNs considering the GNN architecture, optimization techniques, and specific graph structures. For GNN architecture, current works predominantly focus on standard architectures like GCN or MPNNs, often limited to single-layer models or fixed weight matrices that do not take the training process of GNNs into consideration. Future investigations could explore popular GNN designs such as attention mechanisms, skip connections with multiple layers, and learnable weight matrices. In terms of optimization algorithms, while some studies analyze generalization bounds using standard Stochastic Gradient Descent (SGD), the impact of other training techniques like momentum, adaptive learning rates, gradient clipping, and normalization on generalization remains largely unexplored. Another avenue for research involves deriving generalization bounds specific to different types of graphs, such as directed graphs, sparse graphs, heterophily graphs, and dynamic graphs, by imposing additional constraints tailored to each graph type. This approach could lead to tighter generalization bounds that offer insights into the properties of diverse graph structures. Besides, deriving a sharp generalization bound that has a lower bound matching the upper bounds is also promising direction to characterize the generalization ability of GNNs precisely.  

In the preceding discussion, we have scrutinized the generalization capacity of GNNs by examining the generalization boundary in the context of in-distribution generalization, assuming training and testing graph data are drawn from the same distribution. However, real-world scenarios frequently exhibit distribution shifts between training and test data, leading to a notable decline in model performance. Consequently, out-of-distribution(OOD) emerges as a crucial area for evaluating the generalization prowess of GNNs. Despite the proposition and successful implementation of various OOD generalization algorithms with theoretical assurances, systematically reviewing the theory of OOD generalization on graphs poses significant challenges due to the intricate nature of graph-related tasks, diverse distribution shift types such as varying graph sizes and distinct feature distributions, and the evolving GNN frameworks inspired by cross-domain knowledge.  Despite these complexities, notable theoretical advancements have been made in OOD generalization on graphs. Xu et al.~\cite{xu2020neural} investigate the extrapolation capabilities of GNNs trained via gradient descent concerning algorithm alignment within the aforementioned Neural Tangent Kernel (NTK) framework. Ma et al.~\cite{ma2021subgroup} establish the generalization boundary of GNNs for node classification across any subgroup of unlabeled nodes under distribution shift using the Probabilistic Approximate Correctness (PAC)-Bayesian framework. Additionally, Zhou et al.~\cite{zhou2022ood} demonstrate that link prediction based on permutation-equivariant node embeddings obtained through GNNs on graphs of increasing size tends to converge to random guessing, thereby compromising OOD generalization capabilities. For an comprehensive overview of methodologies and strategies pertaining to OOD problem on graphs, we highly recommend readers referring to Li et al.~\cite{li2022out}.

\section{Optimization}
\label{sec:opt}
In previous section, we have discussed the generalization and expressive power but typically neglect the training process to obtain such GNNs, which involves the optimization of GNNs. The goal of optimization in the training process of GNNs is to find the optimal parameters that minimize the loss on training samples, which can be expressed as
\begin{equation}
    \theta = \arg \min_{\theta}L(\theta)\triangleq\frac{1}{n}\sum_{i=1}^{n}\ell\left(f_{\theta}(x_i),y_i\right).\footnote{Here we omit the regularization term for simplicity.}
\end{equation}

The field of optimization theory in deep learning explores the model training procedures, addressing concerns related to the model's convergence towards to optimal solutions and the speed of this convergence. However, compared to the extensive exploration of generalization, expressiveness in GNNs, and their counterparts in deep learning, the study of GNN optimization theory has been rarely explored. This is primarily attributed to the complex training dynamics introduced by graph convolutions and the diverse array of methods and techniques employed to facilitate GNN training. In this section, we will review the theory of optimization of GNNs from three aspects. First, we present the works revealing the dynamics of gradient descent in training GNNs, which is the foundation of optimization. Then, we focus on how the training process of GNNs benefited from some useful training techniques, including weight initialization and normalization. Lastly, we will introduce graph sampling techniques tailored for variance reduction, devised to enhance the efficiency of GNN training processes and bolster the scalability of GNN models.

\subsection{Dynamics of gradient descent in GNN}
Gradient descent is a widely used optimization algorithm in deep learning that updates the parameters following the negative gradient of the loss function w.r.t. to the parameters. Although the gradient descent is popular, the dynamics of gradient descent for training GNNs is understudied due to the non-convexity and non-linearity of the graph convolutional operation with non-linear activation. Since the graph convolutional operation is highly related to graph structure, the dynamics of gradient descent in GNNs can promote the understanding of the role of the graph structure in the training of GNNs. In this subsection, we will review the preliminary attempts to analyze the dynamics of gradient descent in GNNs and most of them follow the optimization theory developed in deep learning. To be specific, they usually consider GNNs in linearized or NTK regime or consider shallow GNNs with only one hidden layer. Besides, we only focus on the basic form of gradient descent that is $\theta_{t+1} = \theta_{t} - \eta\nabla L(\theta_t)$ where $\eta$ is the learning rate.

Xu et al.~\cite{xu2021optimization} study the gradient dynamics of GNNs for the first time via linearized GNN that are GNNs with linear activation while maintaining the non-linear property of the dynamics by utilizing the non-convex loss function. Owing to the highly similar behavior and performance in training linearized and ReLU GNN empirically, the setting is meaningful and can provide insights in understanding the training dynamics of real GNN architectures. Analyzing the gradient dynamics in the form of gradient flow, the authors prove that a multi-layer linearized GNN trained by gradient descent with squared loss converge to its global minimum at a linear rate. The main result is given as follows:

\begin{theorem}[Dynamics of gradient descent in linear GNNs]\label{xu_optimization}
Let $f$ be a $L$-layer linear GNN defined as $f(\mathbf{A},\mathbf{X},\mathbf{W})=\mathbf{W}_q\left[\mathbf{W}_L\left(\ldots\left(\mathbf{W}_2\left(\mathbf{W}_1\mathbf{X}\widehat{\mathbf{A}}\right)\widehat{\mathbf{A}}\right)\ldots\right)\widehat{\mathbf{A}}\right]$. $\mathbf{W}_t$ represents the collection of parameters at time $t>0$ with initialization $\mathbf{W}_0$. $\ell(\mathbf{W}_t)$ denotes the training loss of $f$ with parameters $\mathbf{W}_t$ and $\ell^*$ denotes the global minimum of the training loss. The loss function used here is squared loss. Then, for any $T>0$, we have
\begin{equation}
\ell\left(\mathbf{W}_T\right)-\ell^* \leq\left(\ell\left(\mathbf{W}_0\right)-\ell^*\right) e^{-4 \lambda_T^{(L)} \omega_{\min }^2\left(\mathbf{X}\left(\widehat{\mathbf{A}}^L\right)_{* \mathcal{I}}\right) T},
\end{equation}
where $\lambda_T^{(L)}$ is the smallest singular eigenvalue of the multiplying parameter matrices up to $T$, that is, $\lambda_T^{(L)}:=\inf_{[0,T]}\lambda_{min}((\widehat{\mathbf{W}}_t^{(1:L)})^T\widehat{\mathbf{W}}_t^{(1:L)})$ and $\widehat{\mathbf{W}}^{(1:l)}:=\mathbf{W}_{(l)}\mathbf{W}_{(l-1)}\ldots \mathbf{W}_{(1)}$ for any $l\in\{0,\ldots,L\}$ with $\widehat{\mathbf{W}}_t^{(1:0)}:=I$. $\omega_{min}(\cdot)$ denotes the smallest singular value of the matrix. $(\cdot)_{*\mathcal{I}}$ represents the sub-matrix composed of the columns indexed by the labeled samples.
\end{theorem}

Theorem~\ref{xu_optimization} implies the dependence on several factors and further guarantee the linear convergence rate of linearized GNN to the global minimum as long as $\omega_{\min }^2\left(\mathbf{X}\left(\widehat{{\mathbf{A}}}^L\right)_{* \mathcal{I}}\right)>0$ and $\lambda_T^{(L)}>0$ for $T>0$, which is empirically verified. Besides, the authors further show that the latter condition can be satisfied by proper initialization. 

Different from Xu's work that analyze the gradient dynamics of GNNs in weight space, Yang et al.~\cite{yang2023graph} focus on the evolution of the function learned by GNN with ReLU activation and arbitrary number of layers to demonstrate how GNNs utilize graph structure information during training. To be specific, the authors utilize a node-level GNTK ro prove that the optimization of GNNs actually performs a kernel-graph alignment in NTK regime. Specifically, as proved in Section ~\ref{sec:gen}, when the width goes to infinity the kernel matrix will eventually converge to the deterministic kernel matrix in $t=0$ and be constant during training. Therefore, the GNTK can be viewed as a constant kernel. To perform transformation and propagation step in the form of NTK in each layer, the optimization algorithm incorporates the adjacent matrix into the kernel function in propagation step, which indicates that the gradient descent optimization of GNNs implicitly utilizes the graph structure information to promote training and performs a kernel-graph alignment.

Lin et al.~\cite{lin2023graph} also study the training dynamics of GNNs with ReLU activation and Gaussian initialization in NTK regime and further take the graph information into consideration by introducing a novel measurement named graph disparity coefficient that quantifies the dissimilarity between graph feature and graph structure i.e. graph Laplacian. The authors provide a high-probability convergence guarantee of over-parameterized GNNs trained by gradient descent to demonstrates that the GNNs can converge to its global minimum with high probability as the width of GNNs increases. Besides, the iterations required to achieve global minimum is $O(\tau^2poly(D,L,N))$ where $\tau$,$D$,$L$, and $N$ are graph disparity coefficient, width of GNN, number of layers and number of nodes respectively, which is in accord with the intuition that a small graph disparity coefficient corresponding to a high consistency between graph feature and graph structure can accelerate the convergence.

There are also a few works delve into the training dynamics of GNNs with shallow architectures. Zhang et al.~\cite{zhang2020fast} introduce a learning algorithm characterized by specific tensor initialization and accelerated gradient descent techniques, aiming to facilitate the convergence of one-hidden-layer GNNs to their global minima with zero generalization error. This approach is framed within the context of model estimation, wherein the objective is to reconstruct the parameters of an unknown model sharing an identical architecture. Demonstrated to exhibit linear convergence rates for both regression and binary classification tasks on graphs, this algorithm outperforms conventional gradient descent methods in terms of speed. In a similar vein to Zhang's framework, Awatshi et al.~\cite{awasthi2021convergence} postulate that labels stem from an undisclosed one-hidden-layer GNN, employing a more generalized Gaussian initialization scheme for weight matrices and input features. Noting the unsuitability of the prevalent NTK regime, known for its extremely slow convergence in highly over-parameterized neural networks and inconsistent behavior with real-world GNN training dynamics, the researchers leverage dual activation approaches~\cite{daniely2016toward} to surpass NTK constraints. They establish that a single-hidden-layer message-passing GNN employing ReLU activation, Gaussian initialization, and optimized through gradient descent can converge to an expected loss of $\epsilon$ with respect to the squared loss function in O($\frac{1}{\epsilon^2}log(\frac{1}{\epsilon})$) iterations. Besides gradient descent strategies, Yadati et al.~\cite{yadati2022convex} present a convex programming framework, which offers a verifiable equivalence to the training process of a two-layer GCN with ReLU activation. This connection bridges the chasm between the non-convex optimization characteristic of GNN training with nonlinear activations and the well-established convex optimization paradigms, marking a pioneering integration of these disparate optimization theories.

\subsection{Training tricks}
While the gradient descent algorithm is employed to update GNNs parameters iteratively towards optimal values that minimize the loss function, the practical training process of GNNs can encounter overwhelming challenges like vanishing/exploding gradients and over-smoothing. These issues may impede convergence speed or diminish model performance. As a result, several training strategies have been proposed to mitigate these challenges, some with broad applicability in deep learning and others tailored specifically for graph-related tasks. In this section, we will introduce two categories of training techniques: weight initialization and normalization methods. 

\textbf{Weight initialization.} The weight initialization is crucial to avoid vanishing/exploding gradient problems in deep learning. To achieve the goal, some well-known initialization methods such as Kaiming initialization~\cite{he2015delving}, Xavier initialization~\cite{glorot2010understanding} and, Lecun initialization~\cite{lecun2002efficient} have been devised to regulate variance consistency across layers during both forward and backward propagation. The forward variance $var(x^{(l)})$ and backward variance $var(\frac{\partial Loss}{\partial x^{(l)}})$ for $l$-th layer are computed by the mean of the corresponding variance of each node, that is $var(x_i^{(l)})$ and $var(\frac{\partial Loss}{\partial x_i^{(l)}})$ respectively, where the loss function is standard cross entropy loss. Although initially tailored for fully connected neural networks and CNNs, these approaches surprisingly exhibit efficacy when applied to more intricate graph convolution layers entailing a message-passing scheme and diverse graph structures. Inspired by this unforeseen adaptability, Li et al.~\cite{li2023initialization} delve into analyzing forward and backward variances across layers for message-passing GNNs. They further delineate explicit expressions for these variances by deconstructing the computation graph into distinct message propagation paths and subsequent weight propagation paths. The expressions of $var(x_i^{(l)})$ and $var(\frac{\partial Loss}{\partial h_i^{(l)}})$ for node $i$ in $l$-layer are given as follows. 
\begin{equation}  \operatorname{var}\left(x_i^{(l)}\right)=\left(\frac{\prod_{k_1=0}^{l-1} m_1^{\left(k_1\right)}}{2^l}\right)\left(\prod_{k_2=0}^{l-1} \operatorname{var}\left(\mathbf{W}^{(k_2)}\right)\right)\left(\left[\widetilde{\mathbf{A}}^{(l)} x^0\right]_i^2\right),
\end{equation}
\begin{equation}
    \begin{aligned}
        \operatorname{var}\left(\frac{\partial Loss}{\partial x_i^{(l)}}\right)=&\left(\frac{\prod_{k_1=l+1}^{L-1} m_2^{\left(k_1\right)}\left(C-1\right)}{2^{\left(L-l\right)}N^2C}\right) \\
        &\left(\prod_{k_2=l+1}^{L-1} \operatorname{var}\left(w^{(k_2)}\right)\right)\left(\left[\widetilde{\mathbf{A}}^{L-l} \boldsymbol{1}\right]_i^2\right),
    \end{aligned}
\end{equation}
where $var(w^k)$ is the variance of the distribution from which the weight matrix of $k$-th layer are sampled, $m_1^{(k)},m_2^{(k)}$ are the input or output dimension of the weight matrix of $k$-th layer, $C$ is output dimension of the last layer, $N$ is the number of nodes, $x^0$ is a $N$-dimensional vector obtained by the mean over elements of the input feature of each node, $\widetilde{\mathbf{A}}$ is the normalized adjacency matrix of the graph with self-loops, $[.]_i^2$ is the square of $i$-th element of the vector. 

From the above two equations, it is observed that the newly derived variance expressions for GNNs additionally take the graph structure information and the message-passing scheme into consideration. Besides, the variance of nodes within the same layer differs due to the variable receptive field, which is a significant difference from the setting in FNNs and CNNs that neurons shares the same variance at each layer. Based on the analysis, the authors propose a novel initialization method for GNNs named Virgo that stabilizes the variance of each node across layers during the forward and backward propagation respectively. Empirically, the Virgo initialization surpasses the performance of other initialization methods developed in FNNs and CNNs in most cases when testing on multiple GNN architectures and graph learning datasets.

\textbf{Normalization.} Normalization methods play a vital role in enhancing optimization efficiency and model performance in deep learning practices. These methods typically normalize features, weights, or gradients to impart specific statistical characteristics aligning with diverse objectives, thereby bolstering optimization stability. Following the introduction of BatchNorm~\cite{ioffe2015batch}, numerous normalization techniques have emerged across various domains, with extensive research dedicated to scrutinizing their effectiveness. Nevertheless, minimal attention has been directed towards normalization methods customized for GNNs or evaluating the theoretical suitability of normalization techniques from other domains.

Notably, Cai et al.~\cite{cai2021graphnorm} first compare the performance of three well-known normalization methods in deep learning that are BatchNorm~\cite{ioffe2015batch}, LayerNorm~\cite{ba2016layer} and InstanceNorm~\cite{ulyanov2016instance} when adapting to GNNs. The authors show that the InstanceNorm helps GNN converge faster and achieve better performance than BatchNorm while LayerNorm has little effect. Different from previous analysis that owes the success of these normalization to the scale operation~\cite{bjorck2018understanding}, the authors focus on the shift operation in the normalization process and further prove that the shift operation in InstanceNorm can be view as a preconditioning of graph diffusion matrix to reduce the condition number, thus brings about smoother optimization and accelerate the convergence speed of training. However, the BatchNorm is less effective since different batches of graph data can have varying statistics which fail to approximate the statistics of all samples precisely. Besides, the authors demonstrate that the standard shift operation can degrade the expressive power of the GNN architecture for losing graph structure information such as the degree of nodes. Therefore, they devise a novel normalization method named GraphNorm that automatically adjusts the step of shift operation to preserve the graph structure information. Besides Cai's work, there are some other graph-specific normalization methods aiming to deal with problems in graph representation learning such as over-smoothing and varying graph size~\cite{zhao2019pairnorm,zhou2020towards,chen2022learning,zhou2021understanding,chen2023improving,dwivedi2023benchmarking}. Very Recently, Eliasof et al.~\cite{eliasof2024granola} introduce GRANOLA that adaptively perform normalization on node feature according to the input graph via attaching random feature that we mention in Section~\ref{sec:exp_power}  and then passing through an additional GNN, which not only enhances the performance of GNNs across various graph benchmarks and architectures, but also increases the expressive power of the GNN model.

\subsection{Sampling methods performing variance reduction}
In recent years, training GNNs on large graph datasets has attracted much attention for growing size of graph data. However, original GNNs fail to be applied to large graphs directly since the training process of GNNs requires the adjacency matrix and feature matrix of the entire graph and intermediate node embeddings computed by the exponentially growing receptive field of each node with respect to the depth of GNNs, which consumes much memory and can result in extremely low convergence rate in large graphs. Therefore, it is necessary to utilize sampling for training GNNs efficiently and improving the scalability of GNNs~\cite{huang2021scaling,shilmc}. 

Generally, current research efforts in this domain are typically classified into four categories: node-wise sampling, layer-wise sampling, (sub)graph-wise sampling, and sampling methodologies tailored for heterogeneous graphs. For a more detailed and comprehensive survey of sampling methods in GNNs, readers are recommended to consult Liu et al.~\cite{liu2021sampling}. While in this section, we mainly focus on the different approaches to perform variance reduction in various sampling methods that can guarantee the quality of sampling in GNNs from the theoretical perspective. Minimizing the variance is a widely-used optimization objective for an unbiased sampler since the sampling methods in GNNs only select part of nodes, which inevitably introduce variance and bias that can cause performance degradation and low convergence speed. 

\textbf{Historical activation.}The historical activation method leverages past node embeddings as an approximation of true neighbor node embeddings for variance reduction in aggregation. This approach sidesteps the need for recursively computing neighbor node embeddings, thereby enhancing training efficiency. Chen et al.~\cite{chen2018stochastic} introduce VR-GNN, which initially employs this technique in node-wise sampling as a control variate. They subsequently conduct a variance analysis to compare the variances of the control variate with neighbor sampling techniques utilized in GraphSAGE~\cite{hamilton2017inductive}. 

The expression of these two variances is given in the following equation:
\begin{align}\label{vrgnn}
    &Var[CV_u^{(l)}] = C\sum_{v_1\in \mathcal{N}(u)}\sum_{v_2\in \mathcal{N}(u)}\left(\widehat{\mathbf{A}}_{uv_1}\Delta x_{v_1}^{(l)}-\widehat{\mathbf{A}}_{uv_2}\Delta x_{v_2}^{(l)}\right)^2 \\
    &Var[NS_u^{(l)}] = C \sum_{v_1\in \mathcal{N}(u)}\sum_{v_2\in \mathcal{N}(u)}\left(\widehat{\mathbf{A}}_{uv_1}x_{v_1}^{(l)}-\widehat{\mathbf{A}}_{uv_2}x_{v_2}^{(l)}\right)^2
\end{align}
where $Var[CV_u^{(l)}]$ and $Var[NS_u^{(l)}]$ denotes the variance of control variate and neighbor sampling respectively, $C$ is an constant that depends on the input graph, $\Delta x_v^{(l)} = x_v^{(l)}-\overline{x}_v^{(l)}$ denotes the difference between real activation and historical activation which is small when the parameters of GNNs does not change much. 

Since the $\overline{x}_v^{(l)}$ is often significantly smaller than $x_v^{(l)}$, the historical activation do help reduce the variation. Besides, the authors prove that training GNNs with the control variate by SGD converges to a local minimum with a convergence rate of $\mathcal{O}\left(\frac{1}{\sqrt{T}}\right)$ regardless of sampling, which indicates the variance of the approximation can be zero eventually. Similarly, Cong et al.~\cite{cong2020minimal} utilize historical node embeddings from the preceding layer to diminish variance during forward propagation, employing a layer-wise sampling approach. While the historical activation method proves effective in variance reduction, it necessitates extra memory usage, particularly in node-wise sampling techniques where each node's receptive field continues to grow exponentially. 

\textbf{Importance sampling.} Importance sampling is a commonly employed technique for variance reduction, used to approximate the expectation over distribution $p$ with respect to a distribution $q$ in a Monte Carlo estimator by reweighting samples. This process can be formulated as follows.
\begin{equation}
    \mathbb{E}_p[f(x)]=\int f(x)p(x)dx=\mathbb{E}_q[f(x)\left(\frac{p(x)}{q(x)}\right)],
\end{equation}
where $q(x)$ and $p(x)$ have the same support of distribution and $\frac{p(x)}{q(x)}$ is called the importance function. 

Chen et al.~\cite{chen2018fastgcn} propose Fast-GCN that first utilizes importance sampling in GNNs to reduce the variance by viewing the forward propagation of GNN as integral transforms of embedding function under a sampling distribution, which is given in the following equation:
\begin{equation}\label{integral_gnn}
    x_v^{(l+1)}=\phi\left(\int \widehat{\mathbf{A}}_{vu}x_u^{(l)}(u)\mathbf{W}^{(l)}dP(u)\right),
\end{equation}
where $P$ is the sampling distribution. Then the authors sample a fixed number of nodes in each layer independently to approximate the embedding function under a specific importance distribution. The total average of sample approximation is given in the following equation.
\begin{equation}\label{fastgnn}
    G_{st} = \frac{1}{st}\sum_{i=1}^{s}\sum_{j=1}^{t}\left( \widehat{\mathbf{A}}_{v_iu_j} x_{u_j}^{\prime}\left(\frac{dP(u)}{dQ(u)}\bigg|_{u_j}\right)\right),
\end{equation}
where $s,t$ are the numbers of sampled nodes in the $l+1$-th and $l$-th layer respectively, $v_i,u_j$ are sampled nodes in the $l+1$-th and $l$-th layer respectively and $x_u^{\prime}$ equals to $x_u^{(l)}\mathbf{W}^{(l)}$. To minimize the variance of Equation~\ref{fastgnn}, the optimal distribution $Q$ should satisfy
\begin{equation}
    dQ(u)=\frac{b_u|x_u^{\prime}|dP(u)}{\int b_u|x_u^{\prime}|dP(u)}, 
\end{equation}
where $b_u$ equals to $\left[\int \widehat{\mathbf{A}}_{vu}^2dP(v)\right]^{\frac{1}{2}}$. However, the $|x_u^{\prime}|$ is expensive to compute for changing constantly thus the authors utilize sampling probability proportional to $||\widehat{\mathbf{A}}_{:,u}||^2$ instead that is fixed during training. 

Different from Chen's work that samples nodes in each layer independently, Huang et al.~\cite{huang2018adaptive} adopt a top-down layer-wise sampling manner in which nodes sampled in the lower layer rely on the nodes sampled in the upper layer, which takes the connection between layers into account. The approximation of $x_{v_i}^{(l+1)}$ using adaptive layer-wise importance sampling is given in the following equation: 
\begin{equation}\label{asgnn}    \phi\left(N(v_i)\mathbb{E}_{q(u_j|v_1,\ldots,v_n)}\left[\frac{p(u_j|v_i)}{q(u_j|v_1,\ldots,v_n)}\right]x_{u_j}^{(l)}\mathbf{W}^{(l)}\right),
\end{equation}
where $N(v_i)$ equals to $\sum_{j=1}^{n}\widehat{\mathbf{A}}_{v_iu_j}$, $p(u_j|v_i)$ is the probability of sampling $u_j$ when $v_i$ is given that equals $\frac{\widehat{\mathbf{A}}_{v_iu_j}}{N(v_i)}$, $q(u_j|v_1,v_2,\ldots,v_n)$ is the probability of sampling $u_j$ when all the nodes in the upper layer are given. Then Equation~\ref{asgnn} can be approximated via a Monte-Carlo estimator and the variance is given as follows.
\begin{equation}
    Var = \frac{1}{n^{\prime}}\mathbb{E}_{q(u_j)}\left[\frac{\left(p(u_j|v_i)|x_{u_j}^{(l)}|-\mu_q(v_i)q(u_j)\right)^2}{q^2(u_j)}\right],
\end{equation}
where $q(u_j)$ denotes $q(u_j|v_1,v_2,\ldots,v_n)$, $\mu_q(v_i)$ is the expectation of the estimator. The optimal that minimize the variance is given in the following equation:
\begin{equation}~\label{distribution_asgnn}
    q^*(u_j) = \frac{p(u_j|v_i)|x_{(u_j)}^{(l)}|}{\sum_{j=1}^{n}p(u_j|v_i)|x_{u_j}^{(l)}|}.
\end{equation}

However, the computation of optimal distribution in Equation~\ref{distribution_asgnn} is impractical since $x_{u_j}^{(l)}$ is inaccessible in a top-down sampling manner. Therefore, the authors replace $x_{u_j}^{(l)}$ with a learnable function $g(x(u_j))$ and further add the variance to the loss function to explicitly reduce the variance during training. Alternatively, Liu et al.~\cite{liu2020bandit} propose a bandit sampling method that optimizes the variance from the perspective of the bandit to handle the uncomputable term. The approximation ratio of the variance of the obtained bandit sampler is proved to approach 3 asymptotically. Zou et al.~\cite{zou2019layer} also consider layer-dependent importance sampling in a top-down manner while the sampling is only performed among the union of a neighborhood of nodes in the upper layer to maintain the density of adjacency matrix between layers. 

Instead of sampling nodes in each layer, Zeng et al.~\cite{zeng2019graphsaint} sample subgraphs of the original training graph to build a mini-batch for training GNNs efficiently. To be specific, the subgraphs are constructed via samplers for node and edge respectively with probability distribution derived by importance sampling aiming to minimize the variance. Notably, the optimal sampling probability of an edge $e=(u,v)$ is given by
\begin{equation}
    p_e = \frac{m}{\sum_{e^{\prime}}||\sum_{l}\mathbf{b}_{e^{\prime}}^{(l)}||}||\sum_{l}\mathbf{b}_{e}^{(l)}||
\end{equation}
where $\mathbf{b}_{e}^{(l)}=\widehat{\mathbf{A}}_{vu}x_u^{(l-1)}+\widehat{\mathbf{A}}_{uv}x_v^{(l-1)}$. To alleviate the computation burden, the simplified probability distribution is proportional to $\frac{1}{deg(u)}+\frac{1}{deg(v)}$ that only depends on graph topology, which can be explained by the intuition that two connected nodes with few neighborhoods tend to have a big influence to each other and sampled in the same subgraph with high probability. Different from previous works, Cong et al.~\cite{cong2020minimal} apply a gradient-based adaptive importance sampling to reduce stochastic gradient variance during the training process of GNNs optimized by SGD. To be specific, the authors utilize the estimated norm of gradient and then calculate the importance distribution to sample nodes with minimal variance.

\subsection{Discussion}
In this section, we have reviewed the theory of optimization of GNNs in terms of the dynamics of gradient descent in training GNNs, the training tricks as well as the sampling methods for training GNNs efficiently. Besides the theoretical results, we also introduce some practical methods developed to improve the training process of GNNs. Although GNNs can be properly trained currently, many challenges in delving into the theory of optimization still remain.

One major challenge is to study the dynamics of gradient descent for training general GNNs since the existing works either consider shallow GNNs instead or analyze the optimization of deep GNNs in linearized or NTK regime, which is quite different from the realistic GNN architecture or training behavior of gradient descent. Furthermore, characterizing the dynamics of another optimization algorithm such as SGD and Adam for training GNNs is worth exploring. Besides the GNN architectures and optimization algorithms, it is necessary to incorporate structure information of graphs into the analysis of training process of GNNs. It remains a mystery how the graph structure influence the loss landscape of GNNs exactly. For training tricks in GNNs, extra training tricks can be taken into account and a more careful convergence analysis of different methods to derive the concrete convergence rate is desirable. In addition, since the mentioned training tricks are often utilized jointly during the training process of GNNs, it is meaningful to analyze the interplay of different training tricks or study the effectiveness of applying one individual training trick separately in order to provide a more solid theoretical guarantee. Besides, the connection between the optimization and the other two perspectives that are generalization ability and expressive power of GNNs is not well understood. From this point, some works have studied the generalization bound of GNNs under specific optimization algorithms~\cite{tang2023towards,cong2021provable,esser2021learning} and the extra expressive power obtained from some training tricks such as skip connection a normalization~\cite{eliasof2024granola,cai2021graphnorm,xu2018representation}.

\section{GNNs for long-range and high-order interactions}
\label{sec:exp_long_range}
In practical applications, it is uncommon for researchers to implement GNNs with more than 4 layers, in contrast to that in CNNs where a larger number of layers are typically used to capture long-range dependence. This discrepancy is primarily attributed to the challenges faced by deep GNNs, such as over-smoothing~\cite{li2018deeper} and over-squashing~\cite{alon2020bottleneck}, which results in considerable performance degradation. These issues have attracted significant attention in graph learning community, impeding the effective deployment of GNN architectures. Thus, this section will delve into the theoretical background and proposed solutions to above-mentioned phenomena. 

\subsection{Over-smoothing}

We first introduce the definition of over-smoothing where the features of nodes in a graph tend to converge and lose their distinctions as the number of layers in a GNN increases\cite{li2018deeper}. This convergence diminishes the informative content carried by the nodes' features, resulting in a significant decline in GNN performance.  This observation contradicts the common belief that increasing the number of layers continuously enhances performance, which gains insights into the underlying mechanisms of GNNs and further impedes strategies to overcome the issue for robust GNN models.  

\subsubsection{Measurements}
Currently, the main measurements for quantifying over-smoothing are Dirichlet energy~\cite{cai2006note} and MAD~\cite{chen2020measuring} which are based on the similarity between different nodes. The definitions of the two measurements in the $l$-th layer are given as follows.
Dirichlet energy on graphs:
\begin{equation}
    \mathcal{E}(\mathbf{X}^{(l)})=\frac{1}{N}\sum_{i=1}^{N}\sum_{j\in \mathcal{N}_i}||x_i^{(l)}-x_j^{(l)}||_2^2.
\end{equation}
Mean average distance(MAD) on graphs:
\begin{equation}
    \mu(\mathbf{X}^{(l)})=\frac{1}{N}\sum_{i=1}^N\sum_{j\in \mathcal{N}_i}\left(1-\frac{{x_i^{(l)}}^\top x_j^{(l)}}{||x_i^{(l)}||\;||x_j^{(l)}||}\right).
\end{equation}

According to the above definitions, the Dirichlet energy can be interpreted as the average norm of gradients, while the MAD calculates the average cosine similarity between all node pairs. A recent work by Rusch et al.~\cite{rusch2023survey} introduces a refined definition of over-smoothing that is more rigorous and manageable. According to their definition, over-smoothing is characterized as the exponential decline of node similarity towards zero with an increase in the number of layers and the Dirichlet energy emerges as a superior metric compared to MAD that has been empirically supported. 

\subsubsection{Theory}
The cause of over-smoothing has been illustrated via studying the role of graph convolution layer in transforming node feature. Li et al.~\cite{li2018deeper} provide a explanation of over-smoothing for the first time via demonstrating that the graph convolution is essentially a special form of Laplacian smoothing. To be specific, the graph convolution can be obtained by replacing the the normalized Laplacian $\tilde{\mathbf{D}}^{-1}\tilde{\mathbf{L}}$ with symmetric normalized Laplacian $\tilde{\mathbf{D}}^{-\frac{1}{2}}\tilde{\mathbf{L}}\tilde{\mathbf{D}}^{-\frac{1}{2}}$ in standard Laplacian smoothing $(\mathbf{I}-\tilde{\mathbf{D}}^{-1}\tilde{\mathbf{L}})\mathbf{X}$. From the observation that the Laplacian smoothing updates the feature by aggregation the feature of its neighborhood (including itself) to make the feature within the same cluster similar, they further prove that the embedding of nodes within a connected component will converge to the same value after implementing Laplacian smoothing for many times. 
\begin{theorem}\label{li_oversmoothing} Suppose that a graph has no bipartite component and has k connected $\{C_i\}_{i=1}^{k}$. Let $\mathbf{1}^{(i)}\in \mathbb{R}^n$ be the vector that indicates whether a node is in component $C_i$ i.e. $\mathbf{1}^{(i)}_j=1$ if node j is in component $C_i$ otherwise 0. Then for any $w\in\mathbb{R}^n$ and $\alpha\in(0,1]$,
\begin{align}
    &\lim_{m \to \infty}(\mathbf{I}-\alpha \mathbf{D}^{-1}\mathbf{L})^m w= [\mathbf{1}^{(1)}, \mathbf{1}^{(2)}, \ldots, 
\mathbf{1}^{(k)}]\theta_1, \\
    &\lim_{m \to \infty}(\mathbf{I}-\alpha \mathbf{D}^{-\frac{1}{2}}\mathbf{L}\mathbf{D}^{-\frac{1}{2}})^m w= \mathbf{D}^{-\frac{1}{2}}[\mathbf{1}^{(1)}, \mathbf{1}^{(2)}, \ldots, \mathbf{1}^{(k)}]\theta_2,
\end{align}
where $\theta_1\in \mathbb{R}^k, \theta_2\in \mathbb{R}^k$.
\end{theorem}

Theorem~\ref{li_oversmoothing} can be directly applied to graphs with self-loop that have no bipartite component, which implies that the features do converge to linear combination of $\{\mathbf{1}\}_{i=1}^{k}$ or $\{ D^{-\frac{1}{2}}\mathbf{1}\}_{i=1}^{k}$ thus become indistinguishable and cause over-smoothing. Different from Li's work, Xu et al.~\cite{xu2018representation} connect the influence distribution of a node spread by message passing scheme to random walk~\cite{lovasz1993random}. Since the random walk distribution will ultimately converge to its limit distribution that only depends on the graph structure, the representation of different nodes after multiple GCN layers will carry little local information that results in over-smoothing.

On the other hand, Oono and Suzuki~\cite{oono2019graph} represent the propagation of GCN as a dynamical system to characterize the asymptotic behavior of GCNs as the number of layers goes to infinity. The authors provide the following theorem to elaborate over-smoothing.
\begin{theorem}\label{expoential} Consider an undirected graph augmented with self-loops, for any input feature $\mathbf{X}^{(0)}$ of the graph, the output of a l-layer non-linear GCN activated by ReLU $\mathbf{X}^{(l)}$ satisfies $d_{\mathcal{M}}(\mathbf{X}^{(l)})\leq(s\lambda)^l d_{\mathcal{M}}(\mathbf{X}^{(0)})$, where $s$ is the upper bound of the maximum singular value of weight matrix, $\lambda$ is the largest absolute value of the non-one eigenvalue of the augmented normalized adjacency matrix, and $d_{\mathcal{M}}(\mathbf{X})$ is the distance between $\mathbf{X}$ and an invariant space that corresponds to eigenspace associated with the eigenvalue 1. In particular, the $d_{\mathcal{M}}(\mathbf{X}^{(l)})$ exponentially converges to 0 if $s\lambda<1$.
\end{theorem}
The Theorem~\ref{expoential} shows that the output of GCN will exponentially converge an invariant space if the weight satisfies the condition determined by the augmented normalized adjacency matrix(or augmented normalized Laplacian). Since the invariant space is the subspace spanned by the eigenvector corresponding to eigenvalue 1 that only carry information of the connected components and node degrees, the GCN fails to distinguish nodes with different degrees within the same connected component thus suffers from over-smoothing. It is surprising to find that the result is essentially identical to that of linearized GNNs~\cite{li2018deeper,zhao2019pairnorm} which indicates that non-linear activation ReLU is independent of over-smoothing. Besides, from the perspective of graph signal processing, the invariant space corresponding to the lowest frequency of the graph Laplacian agrees with the statement in NT~\cite{nt2019revisiting} that the graph convolution layer is essentially a low-pass filter.

\subsubsection{Solutions}
In this subsection, we briefly introduce the methods proposed for alleviating and overcoming the over-smoothing.

\textbf{Normalization and regularization.} Normalization and regularization are methods that directly based on the definition and measurement of over-smoothing to handle the problem~\cite{zhou2021dirichlet,zhao2019pairnorm,zhou2021understanding,zhou2020towards,chen2020measuring,oono2019graph}. They all impose additional constraints obtained by the measurements on GNNs, but normalization method achieve the goal by normalizing the feature embedding and regularization method satisfy them via regularization. For example, NodeNorm~\cite{zhou2021understanding} normalizes the feature to keep the total pairwise squared distance(TPSD) a constant across every layer while Zhou et al.~\cite{zhou2021dirichlet} regularize the Dirichlet energy within a suitable range for each layer.

\textbf{Skip connection.} Skip connection~\cite{he2016deep} is found effective in solving the problem of gradient explosion and gradient vanishing in deep CNNs. Motivated by it, some works add skip connection to GNNs as an attempt to alleviate oversmoothing in deep GNNs~\cite{klicpera2018predict,chen2020simple,xu2018representation,liu2020towards,luan2019break}. They preserve fraction of initial and intermediate feature in the final embedding during the neighborhood aggregation. For example, JKnet~\cite{xu2018representation} combines all previous embedding in the last layer while GCNII~\cite{chen2020simple} conducts skip connection in each layer. Intuitively, the feature in shallow layers is more distinguishable thus the method is reasonable. 

\textbf{Physics inspired equations.} Recently, some works resort to ordinary differential equations(ODEs) or partial differential equations(PDEs) derived from physics to copy with over-smoothing~\cite{rusch2022graph,eliasof2021pde,wang2022acmp,di2022graph,chamberlain2021grand,bodnar2022neural,chamberlain2021beltrami,maskey2024fractional}. To be specific, yhey utilize the physical equations such as diffusion and gradient flows to represent a distinct dynamics beyond original message passing scheme on graphs and then solve them by discretizing the equations to generate novel GNN architectures that are more powerful to solve over-smoothing. For example, Rusch et al.~\cite{rusch2022graph} propose GraphCON based on graph-coupled oscillator and prove that the zero-Dirichlet energy steady states are not stable in the system, which prevent form over-smoothing.

\textbf{Graph rewiring.} Graph rewiring serves as a method to mitigate over-smoothing by adjusting graph topologies. DropEdge~\cite{rong2019dropedge} is a straightforward technique for graph rewiring. By randomly removing edges and decreasing node connections, this approach alleviates over-smoothing effects. Additionally, the authors offer a theoretical insight into the technique, demonstrating that DropEdge can decelerate the convergence rate of GNNs towards the constant space outlined in the previous section, thereby reducing information loss. Besides, Chen et al.~\cite{chen2020measuring} and Hasanzadeh et al.~\cite{hasanzadeh2020bayesian} propose strategies to modify the graph topology adaptively. 

\subsection{Over-squashing}
In contrast to over-smoothing, which hinders the performance of deep GNNs, over-squashing presents challenges in effectively learning long-range interactions. Specifically, as the number of GNN layers increases, the receptive field of a node expands exponentially. This leads to an excessive amount of information being compressed into a fixed-length feature vector, resulting in an information bottleneck known as over-squashing. Consequently, during message passing, long-range information gets distorted, impacting the performance of GNNs on tasks involving large graphs and long-range dependencies.

\textbf{Measurements.} One direct measurement to quantify over-squashing is Jacobian matrix $\frac{\partial x_j^{(d)}}{\partial x_i}$ that analyze the impact on the feature of node $j$ by input feature of node $i$ at distant $r$~\cite{topping2021understanding,bober2022rewiring}. Topping et al.~\cite{topping2021understanding} further proves an upper bound of the Jacobian in the following equation:
\begin{equation}\label{jacobian}
    \left|\frac{\partial x_j^{(r+1)}}{\partial x_i}\right|\leq (\alpha_1\alpha_2)^{r+1}(\widehat{\mathbf{A}}^{r+1})_{ji},
\end{equation}
where $\alpha_1$ and $\alpha_2$ are upper bound of the norm of the gradient of update function and aggregation function respectively. In Equation~\ref{jacobian}, the norm of Jacobian matrix is controlled by the corresponding entry of the power of the augmented normalized adjacent matrix, which suggests that the information from distant node decays exponentially during message passing thus result in over-squashing. Later, the result is generalized to any pair of nodes and the norm of Jacobian is bounded more precisely~\cite{black2023understanding}. 

\textbf{Solutions.} 
The above discussion highlights the importance of amplifying information flow from distant nodes to alleviate over-squashing in graphs and the aforementioned graph rewiring is an effective to achieve the goal. There are wo main types of graph rewiring methods that are spatial graph rewiring~\cite{alon2020bottleneck,topping2021understanding,giraldo2022understanding,nguyen2023revisiting,fesser2024mitigating,liu2023curvdrop,shi2023curvature,sanders2023curvature} and spectral graph rewiring ~\cite{arnaiz2022diffwire,deac2022expander,banerjee2022oversquashing,black2023understanding,karhadkar2022fosr,barbero2023locality,gutteridge2023drew}. The spatial graph rewiring usually connect one node to another node within its receptive field while spectral graph rewiring often optimize some metrics that measures connectivity of graphs. Therefore, the spatial graph rewiring methods modify the edges in a more local manner than spectral graph rewiring methods. For example, Topping et al.~\cite{topping2021understanding} propose Balanced Forman curvature of edges to quantify the over-squashing between nodes at distance 2 and further address the negatively curved edges that are susceptible to over-squashing. On the other hand, the quantities for spectral graph rewiring methods to optimize include spectral gap~\cite{banerjee2022oversquashing}, commute time~\cite{di2023over} and total effective resistance~\cite{black2023understanding}, which are defined globally and closely related to the well-known Cheeger constant in measuring connectedness of the whole graph. It is also worth noting that the graph transformer~\cite{ying2021transformers,wu2023difformer,he2023generalization} can be viewed as an extreme case of spatial graph rewiring for considering fully-connected graphs. However, the interplay between these two categories and the theoretical superiority of one over the other remain unexplored, making it a promising direction for future research.

Besides the graph rewiring methods, there are some other methods to capture the long-range dependencies between node thus alleviate over-squashing. One effective method is the physics inspired GNNs mentioned in over-smoothing that changes the dynamics of message passing~\cite{chamberlain2021grand,chamberlain2021beltrami,toth2022capturing,maskey2024fractional}. Since the method changes the information flow and alters the connectivity of the graphs essentially, they actually perform graph rewiring implicitly. In addition, other methods handle over-squashing problem from their own perspective such as graph imbalance learning~\cite{sun2022position}, expressive powerful~\cite{beaini2021directional} and reservoir computing model that is training-free~\cite{tortorella2022leave}.  For a more detailed and comprehensive survey of over-squashing, we highly recommend readers refer to Akansha et al.~\cite{akansha2023over} and Shi et al.~\cite{shi2023exposition}.
\subsection{Discussion}
In this section, we have reviewed the of theory and solutions of over-smoothing and over-squashing phenomena in GNNs, which prevent the GNN architectures from going deeper and further capturing high-order and long-range interactions between nodes. Next, we outline some open questions for over-smoothing and over-squashing.

\label{sub_sec:smoothing_squashing}
\textbf{Trade-offs between over-smoothing and over-squashing}. In recently years, an increasing number of works based on graph rewiring focus on handling both over-smoothing and over-squashing~\cite{karhadkar2022fosr,giraldo2022understanding,liu2023curvdrop,nguyen2023revisiting} and the tradeoffs between between over-smoothing and over-squashing is emphasized. Intuitively, the graph rewiring methods to alleviate over-squashing often introduce additional edges to original graph, which exerts a smoothing effect on the graph thus poses a risk of over-smoothing. To provide a more theoretical explanation, the tradeoffs can be analyzed from the perspectives of spatial and spectral methods. Nguyen et al.~\cite{nguyen2018continuous} utilize Ollivier-Ricci curvature to establish a geometric connection between over-smoothing and over-squashing in a unified framework. To be specific, the positive graph curvature is associated to over-smoothing while the positive graph curvature is related to over-squashing. Different from Nguyen's analysis that is based on the theory of spatial method, Giraldo et al.~\cite{giraldo2022understanding} study the spectral gap of linearized GNNs and reveal the relationship between over-smoothing and over-squashing. Besides, it is noted that the tradeoffs between over-smoothing and over-squashing can be viewed as a compromise between locality and connectivity of the graph, which provides a novel perspective to explore the interplay of spatial and spectral graph rewiring methods.

\textbf{Connection with heterophily.} The heterophily problem has drawn much attention recently since the majority of GNNs following homophily assumption that nodes with similar feature or identical labels tend to connect each other fail on heterophilic graphs. Although the heterophily and over-smoothing seem to be two independent problem, some works aimed to solve over-smoothing also perform well on heterophilic graphs empirically~\cite{chen2020simple,wang2022acmp} and vice versa~\cite{bo2021beyond,yang2021diverse,liu2021non}. Therefore, it remains an open question that whether the two problem can be solved simultaneously from a theoretical perspective. Yan et al.~\cite{yan2022two} establish the connection between over-smoothing and hetetephily problem for the first time by studying the behavior of linear SGC on random graphs. Specifically, they analyze the change of node representation after message passing in terms of two quantities called relative degree and homophily level respectively. Parallel to Yan's work, Bodnar et al.~\cite{bodnar2022neural} utilize heat diffusion PDE to explain the susceptibility of GNNs to over-smoothing and heterophily. Furthermore, the authors analyze the problems more precisely from a topological perspective based on (cellular) sheaf theory~\cite{bredon2012sheaf}. Besides, it is noted that both of the works prove that GNNs with signed message can handle the two problems simultaneously.

\section{Conclusion}
\label{sec:conclusion}
This survey attempts to furnish a comprehensive overview of theoretical foundations and advancements in graph learning. Given its status as a vibrant research domain intertwined with a diverse array of mathematical linkages, it proves unfeasible to encompass all existing works within the scope of this study. The selections covers three main topics, namely: expressiveness power, generalization ability, optimization techniques. Additionally, the long-range and high-order interaction of GNNs that are populaar topic in recently are also elaborated. In each section, we introduce necessitate preliminaries, systematically elaborate the theoretical findings, and discuss the limitations as well as future directions. In particular, our approach tailors the exposition of each topic with respect to its developmental vein and orientation. 

The expressive power of GNNs often correlates with their ability to distinguish non-isomorphic graphs. We first delve into the theoretical connects GNNs with the extensively studied Weisfeiler-Lehman (WL) algorithm, exploring various research frontiers like strategies to transcend the constraints of the 1-WL and discussing novel architectures such as graph transformers and geometric GNNs. Regarding generalization, the literature is structured around the tools utilized for deriving the generalization bound, elucidating pivotal findings and insights, outlining constraints, and highlighting emerging patterns. In the realm of algorithm optimization processes, we first discuss gradient dynamics of GNNs and present theoretical analyses of several training tricks as well as sampling methods. Despite significant research efforts and notable strides forward, persistent challenges persist in the theoretical process of graph models, notably stemming from the intricate nature of inter-node relationships and the convolutional layers or units inherent to models for graph-structured data. Analytical simplifications applied to either graph properties or model architectures lead to findings that may lack practical relevance across diverse real-world applications.

While expressiveness, generalization, and optimization are conventionally addressed in isolation, there is increasing interest in exploring their interplay to enhance the efficacy of graph neural networks. Key questions arise, such as how optimization strategies like gradient descent and skip connections influence both generalization and expressive power. Additionally, understanding the interconnections between generalization capacity and expressive power, particularly in the context of high-order GNNs~\cite{morris2023wl}, remains a compelling area for further investigation. Besides, since the graphs in real-world have various forms and can be extremely complex, the theoretical results that are able to reflect the intricate graph structure and intrinsic property of graphs are expected. Against the backdrop of groundbreaking advancements in foundational models of computer vision and natural language processing, establishing theoretical frameworks for leveraging large language models on graphs and potentially formulating graph-based foundational models emerges as a critical research frontier. By embracing the challenges and opportunities that lie ahead, we believe more research endeavors will make for continued advancements and transformative impact for graph learning community.


%

\ifCLASSOPTIONcaptionsoff
  \newpage
\fi


\bibliographystyle{IEEEtran}
\bibliography{IEEEabrv,reference}

\end{document}